\documentclass[11pt]{article}

\usepackage[final]{acl}

\usepackage{times}
\usepackage{latexsym}
\usepackage{enumitem}
\usepackage{booktabs}
\usepackage{siunitx}
\usepackage{amsmath}
\usepackage{txfonts}
\usepackage{multirow}
\usepackage{tabularx}

\usepackage[T1]{fontenc}

\usepackage[utf8]{inputenc}

\usepackage{microtype}

\usepackage{inconsolata}

\usepackage{graphicx}

\def\tv{t} 
\def\wv{w}    
\def\iv{i}    

\def\iv{i}

\newcommand{\prob}[1]{\mathsf{P}( #1 )}
\newcommand{\surp}[1]{\mathsf{S}( #1 )}
\newcommand{\condprob}[2]{\mathsf{P}( #1 \mid #2 )}

\title{Clozing the Gap: Exploring Why Language Model Surprisal Outperforms Cloze Surprisal}

\author{Sathvik Nair\footnotemark[1] \\
  University of Maryland \\
  \texttt{sathvik@umd.edu} \\\And
  Byung-Doh Oh\footnotemark[1] \\
  Nanyang Technological University \\
  \texttt{byungdoh.oh@ntu.edu.sg} \\}

\begin{document}
\maketitle
\def\thefootnote{*}\footnotetext[1]{These authors contributed equally to this work.}\def\thefootnote{\arabic{footnote}}
\begin{abstract}
How predictable a word is can be generally quantified in two ways:~using human responses to the cloze task or using probabilities from language models (LMs).
When used as predictors of processing effort, LM probabilities outperform probabilities derived from cloze data.
However, it is important to establish that LM probabilities do so for the right reasons, since different predictors can lead to different scientific conclusions about the role of prediction in language comprehension.
We present evidence for three hypotheses about the apparent advantage of LM probabilities:~not suffering from low resolution, distinguishing semantically similar words, and accurately assigning probabilities to low-frequency words.
These results call for efforts to improve the resolution of cloze studies, coupled with experiments on whether human-like prediction is also as sensitive to the fine-grained distinctions made by LM probabilities.
\end{abstract}

\section{Introduction}

\noindent Prediction is a key principle of language comprehension: Words that are more expected given a context are easier to process compared to unexpected words \citep{ehrlich_1981_contextual, kutas_1984_brain, smith_2013_effect}.
Studying this process requires a measure of how predictable a word is in its context to typical human readers (i.e.~its \textit{predictability}).
The effects of predictability on processing effort have been widely studied and characterized quantitatively \citep{smith_2013_effect, brothers_2021_word, szewczyk_context-based_2022, shain_2024_large}.

\begin{figure}[t!]
    \centering
    \includegraphics[width=0.95\linewidth]{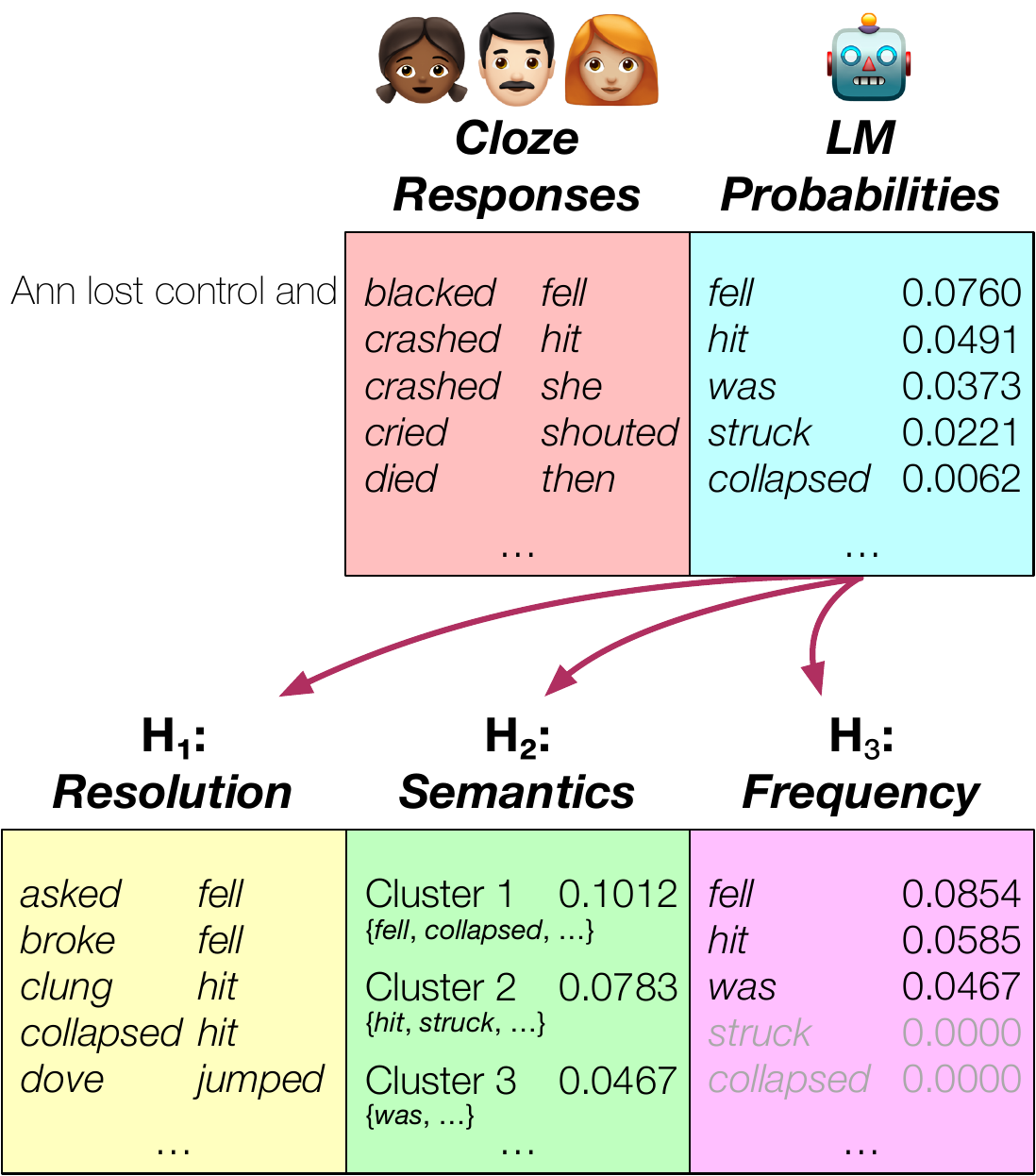}
    \caption{How predictable a word is in its context has traditionally been quantified using the cloze task (red), which is increasingly being replaced with LM probabilities in recent years (blue). After establishing that LM probabilities generally provide a better predictor of reading times than cloze responses (Experiment 1), we test three hypotheses about why (yellow, green, magenta) by conducting a hypothesis-driven manipulation of LM probabilities (Experiment 2).}
    \label{fig:schematic}
\end{figure}

The traditional method for estimating such predictability is the cloze task, where participants provide a completion of the next word given an incomplete sentence \citep{taylor_1953_cloze}: Words that are produced more frequently are considered to be more predictable.
Predictability can also be estimated with language models (LMs): Modern LMs are based on neural networks \citep{radford_2019_language, biderman_2023_pythia, grattafiori_2024_llama} that are trained on a cloze-like task of predicting the next word over large-scale examples of human language use, thus providing accurate estimates of words' conditional probabilities.

Recent experiments show that LM surprisal (negative log probability) generally seems to be a superior predictor of comprehension difficulty compared to cloze surprisal \citep[][though see \citealp{devarda_2024_cloze}]{hofmann_2022_language, michaelov_2023_cloze, shain_2024_large}.
This stronger fit between LM surprisal and human subject data has led some to claim that LMs should replace the cloze task entirely \citep{michaelov_2023_cloze, shain_2024_large}.

However, this claim based on fit to data ignores systematic differences between LM and cloze-based estimates of predictability.
LMs reflect knowledge of billions of training examples and perfect access to the input text, which enable them to make predictions that average readers cannot \citep{oh_2023_surprisal, oh_2025_model}.
Additionally, LM-based estimates of predictability are based \textit{purely} on probabilistic inference and are not explicitly linked to any notion of structure building \citep{slaats_2025_surprising, nair_2026_across}, which limits their ability to model the full range of language processing behavior \citep{stanojevic2023modeling, staub2025predictability}.
On the other hand, cloze data directly come from human participants, but the cloze task is an offline task that elicits a single response from each participant, which may evoke different cognitive processes from those that underlie the graded predictions that are made rapidly \citep{staub_2015_influence, brothers_2023_multiple}, and thus do not fully represent human probabilistic intuitions \citep{smith_2011_cloze}.
This difference appears to underlie the qualitatively different next-word predictions---and therefore predictability estimates---from humans and LMs \citep{jacobs_2024_large, shlegeris_2024_language}.

Crucially, different methods of estimating predictability can lead to different scientific conclusions.
For example, the conclusion that the effect of predictability is better thought of as a \textit{facilitation} at predictable words \citep[instead of a \textit{cost} at unpredictable words;][]{brothers_2021_word} changes when GPT2 probabilities are used instead of cloze probabilities to model exactly the same reading time data \citep{shain_2024_large}.
Even within the same class of LMs, how much the slowdown observed at low-frequency words can be reduced to the effect of predictability changes depending on the specific model used to calculate predictability estimates \citep{oh_2025_dissociable}.
Therefore, we argue that it is important to establish where the relative contribution of cloze and LM surprisal in modeling human data comes from, instead of advocating for the use of LM surprisal purely on the basis of stronger fit to data.

To this end, we first re-establish LM surprisal as a superior predictor of reading times compared to cloze surprisal, while additionally exploring how cloze probabilities should be smoothed and transformed.
We then adopt a novel approach of intervening on the LM's probabilities to test three hypotheses that may account for the stronger fit of LM surprisal (Figure \ref{fig:schematic}).
We find support for all three of our hypotheses: LMs appear to yield stronger predictors of RTs than cloze responses due to their ability to assign highly fine-grained probabilities, distinguish between semantically similar words, as well as between low-frequency words.
This calls for efforts to improve the resolution of cloze studies and for experiments on whether humans' predictions are as sensitive to the fine-grained distinctions made by LM probabilities.\footnote{The code used in this work is available at \url{https://github.com/sathvikn/cloze-surprisal}.}

\section{Experiment 1: Evaluation of Cloze and LM Surprisal as Predictors of RTs} \label{sec:exp1}
The first experiment compares the fit of cloze probabilities and LM probabilities to by-word reading times (RTs).
Using multiple English reading time datasets for which cloze responses are available, we fit regression models that contain either cloze or LM probability as a predictor.
We then compare the fit of these models against a regression model that contains both predictors to evaluate whether one predictor subsumes the effect of the other.
In order to systematically compare the two measures across datasets, we place an increased emphasis on experimenting with different methods for smoothing and transforming cloze probabilities.

\subsection{Response Data: By-Word Reading Times}
\label{sec:corpora}
We analyze four English reading time datasets with available cloze response data.
The stimuli text, cloze responses, and reading times of each dataset are as follows:

\paragraph{BK21 SPR \textnormal{\citep{brothers_2021_word}}.}
The BK21 self-paced reading (SPR) dataset contains 216 triplets of sentences that manipulate the preceding context to make the same target word highly predictable (high-cloze), moderately predictable (moderate-cloze), and rarely predictable (low-cloze).
The items were validated through a cloze norming study that resulted in about 90 responses per sentence: For the high-, moderate-, and low-cloze conditions, the subjects predicted the correct target word around 91\%, 20\%, and 1\% of the time respectively.
We present the same example triplet from \citet{brothers_2021_word}:
\begin{itemize}[leftmargin=*, itemsep=0em]
    \item High-cloze: Her vision is terrible and she has to wear \underline{\textbf{glasses} in class}.
    \item Moderate-cloze: She looks very different when she has to wear \underline{\textbf{glasses} in class}.
    \item Low-cloze: Her mother was adamant that she has to wear \underline{\textbf{glasses} in class}.
\end{itemize}

The proportion of times the target word (i.e.~\underline{\textbf{glasses}}) was produced in the cloze norming study (cf.~the full set of responses) is made publicly available.
This dataset also contains SPR times collected from 216 subjects who did not take part in the cloze norming study, summed over the target word and two succeeding words (i.e.~\underline{\textbf{glasses} in class}).

\paragraph{Provo ET \textnormal{\citep{luke_2018_provo}}.}
The Provo eye-tracking (ET) dataset contains 55 short English paragraphs that are about 50 words long, extracted from news articles, science magazines, and works of fiction (total 2,746 words).
A total of 478 subjects provided cloze responses to each word in the corpus, resulting in about 40 responses for every word.
A separate set of 84 subjects who did not participate in the cloze norming study provided eye-tracking-while-reading data for these paragraphs.
From the raw eye fixation data, we derive and analyze two by-word measures; first-pass (FP) duration, which is the time taken between entering a word region from the left (in the case of English) and exiting it to either the left or right, and go-past (GP) duration, which is the time taken between entering a word region from the left and exiting it to the right, including all regressive fixations.

\paragraph{UCL SPR/ET \textnormal{\citep{frank_2013_reading, devarda_2024_cloze}}.}
The UCL SPR and ET datasets contain 361 isolated sentences that were extracted from online novels written by aspiring authors.
The sentences were chosen such that they consist fully of frequent English words and could readily be interpreted without the surrounding context or any extra-linguistic knowledge.
All 361 sentences were used to collect SPR times from 117 subjects, and a subset of 205 sentences that fit within a single line was used to collect eye-tracking data from 48 subjects \citep{frank_2013_reading}.
In a separate study by \citet{devarda_2024_cloze}, around 80 cloze responses were collected for each word in the smaller subset of 205 sentences.
We analyze the SPR times, FP durations, and GP durations provided as part of these datasets.

\paragraph{Data filtering.} The reading times in each dataset were filtered in standard ways prior to analysis: Reading times of the first and last word of each sentence (all datasets) and each line (ET datasets) were removed to avoid wrap-up effects.
SPR times and GP durations that exceed 3000 ms and FP durations that exceed 2000 ms were also removed.
Finally, for the UCL dataset that also provides sentence-level comprehension data, reading times from trials with incorrect responses were removed.

\subsection{Predictors: Cloze and LM Probabilities}
The main predictors of interest are predictability estimates derived from cloze responses and those calculated from an LM:

\paragraph{Cloze probability smoothing and transform.} Partly due to the limitations in the number of cloze responses (ranging from around $\sim$40 to $\sim$90 for our RT datasets), the actual next word is often not produced, leaving their probabilities to be zero.
This is problematic for the log transform required for surprisal calculation, and therefore the cloze probabilities need to be smoothed by allocating a small probability to zero-count words.

Additionally, the functional form between cloze probability and RTs has not been widely established in the literature.
The linearity of cloze probability on RTs across the high/moderate/low-cloze conditions is reported in \citet{brothers_2021_word}, but it remains to be seen whether this generalizes to other broad-coverage datasets \citep[cf.~the linearity of LM surprisal on RT observed across multiple datasets and languages;][]{wilcox_2023_testing, xu_2023_linearity, shain_2024_large}.

To establish a strong predictor based on cloze responses, we first experiment with different smoothing factors and different functional forms to RTs.
More specifically, we adopt a form of add-one smoothing as follows:

\begin{equation}
\label{eqn:cloze}
    \prob{\wv_\tv} = \frac{C_{\wv_\tv}+1}{\sum_{\wv_\iv} C_{\wv_\iv}+V},
\end{equation}
where $C_{\wv_\iv}$ is the number of times the word $\wv_\iv$ was attested as cloze responses.
In order to control the probability mass that is assigned to unattested cloze completions, we manipulate the smoothing factor $V \in \{50, 100, 200, 500, 1000, 2000\}$.
We also evaluate different functional forms between these cloze probabilities and RTs by transforming them prior to including them in linear mixed-effects regression \citep{xu_2023_linearity}.
The functional forms we experiment with are raw probability $\prob{\wv_\tv}$, raw surprisal $\surp{\wv_\tv} = -\log_{2}\prob{\wv_\tv}$, and various power transforms on surprisal, namely $\surp{\wv_\tv}^{\frac{1}{2}}$, $\surp{\wv_\tv}^{\frac{3}{4}}$, $\surp{\wv_\tv}^{\frac{4}{3}}$, and $\surp{\wv_\tv}^{2}$.

\paragraph{LM surprisal calculation.} The surprisal of each word was calculated from LM-based predictability estimates.
For our LM, we used GPT2 \citep[small;][]{radford_2019_language}, based on its widespread use for next-word probabilities in psycholinguistic modeling studies \citep{huang_2024_large, giulianelli_2026_incremental, kuribayashi_2025_large}.
Each sentence of the BK21 and UCL datasets and each paragraph of the Provo dataset was tokenized and provided to GPT2 to calculate conditional probabilities.
The probability of the directly following whitespace was included as part of the word probability \citep{oh_2024_leading, pimentel_2024_compute} to ensure a proper probability distribution over words.\footnote{Without this correction, for example, if \texttt{carpet} is tokenized into subword tokens \texttt{car} and \texttt{pet}, it always holds that $\prob{\texttt{carpet}} \leq \prob{\texttt{car}}$.}

\subsection{Modeling: LME Regression}
\label{sec:lme}
The ability of cloze probability and LM surprisal to predict reading times is evaluated by their contribution to the log likelihood of a linear-mixed effects \citep[LME;][]{bates_2015_fitting} regression model.
For each of the six measures (BK21 SPR, Provo FP/GP, UCL SPR/FP/GP), we first specify a baseline LME model that includes a set of baseline predictors.
These baseline predictors are word length in characters, position of the word within each sentence, unigram surprisal as a measure of frequency (all measures), and whether or not the preceding word was fixated (FP/GP measures).
Unigram surprisal is calculated using the KenLM toolkit \citep{heafield_2013_scalable} with parameters estimated on $\sim$6.5 billion words of the OpenWebText Corpus \citep{gokaslan_2019_openwebtext}.

To first determine the smoothing and transform of cloze probability that achieves the strongest fit to RTs, we evaluate how much each variant of cloze probability improves the in-sample log likelihood over the baseline LME model.
To this end, we fit a set of 36 LME models (6 smoothing factors $\times$ 6 transforms) that include the baseline predictors and a variant of cloze probability to about 50\% of the observations for each of the six measures.

After choosing the best-fitting variant of cloze probability, the fit of cloze probability is directly compared against that of GPT2 surprisal by comparing nested LME models.
Three LME models are fit to each measure; on top of the baseline model, one additionally containing cloze probability, one additionally containing GPT2 surprisal, and one additionally containing both cloze probability and GPT2 surprisal.\footnote{While it is common to include surprisal from preceding words as predictors of the current word's RT to account for `spillover' effects in reading \citep{vasishth_2006_proper}, we do not include such spillover predictors for two reasons. The first is that for the BK21 dataset, cloze response data is only available at the target word and not for any other word of the sentence, making it impossible to define spillover predictors for cloze probability. More fundamentally, including spillover predictors may introduce a confound for answering our research question; if an LME model containing multiple versions of cloze probability is compared against that containing multiple versions of LM surprisal, it remains unclear whether the observed modeling benefit is attributable to the main predictor or spillover predictors.}
Their fit to RTs was evaluated using 10-fold cross-validation; after splitting each measure into 10 folds,\footnote{The partitioning is conducted by cycling through each subject-by-sentence combination and assigning observations from that combination to a different partition. See Appendix \ref{app:observations} for the number of observations in each fold and their total for each RT measure.} held-out log likelihood is calculated on the fold that was not used for model fitting.
The two comparisons of interest are between the model containing both predictors against that containing only cloze probability and that containing only GPT2 surprisal.
Statistical significance is determined by a paired permutation test over the two sets of 10 per-observation log likelihood values.
All LME models include a by-subject random intercept; the datasets did not consistently support a richer random-effects structure.

\begin{figure}[t!]
    \centering
    \includegraphics[width=\linewidth]{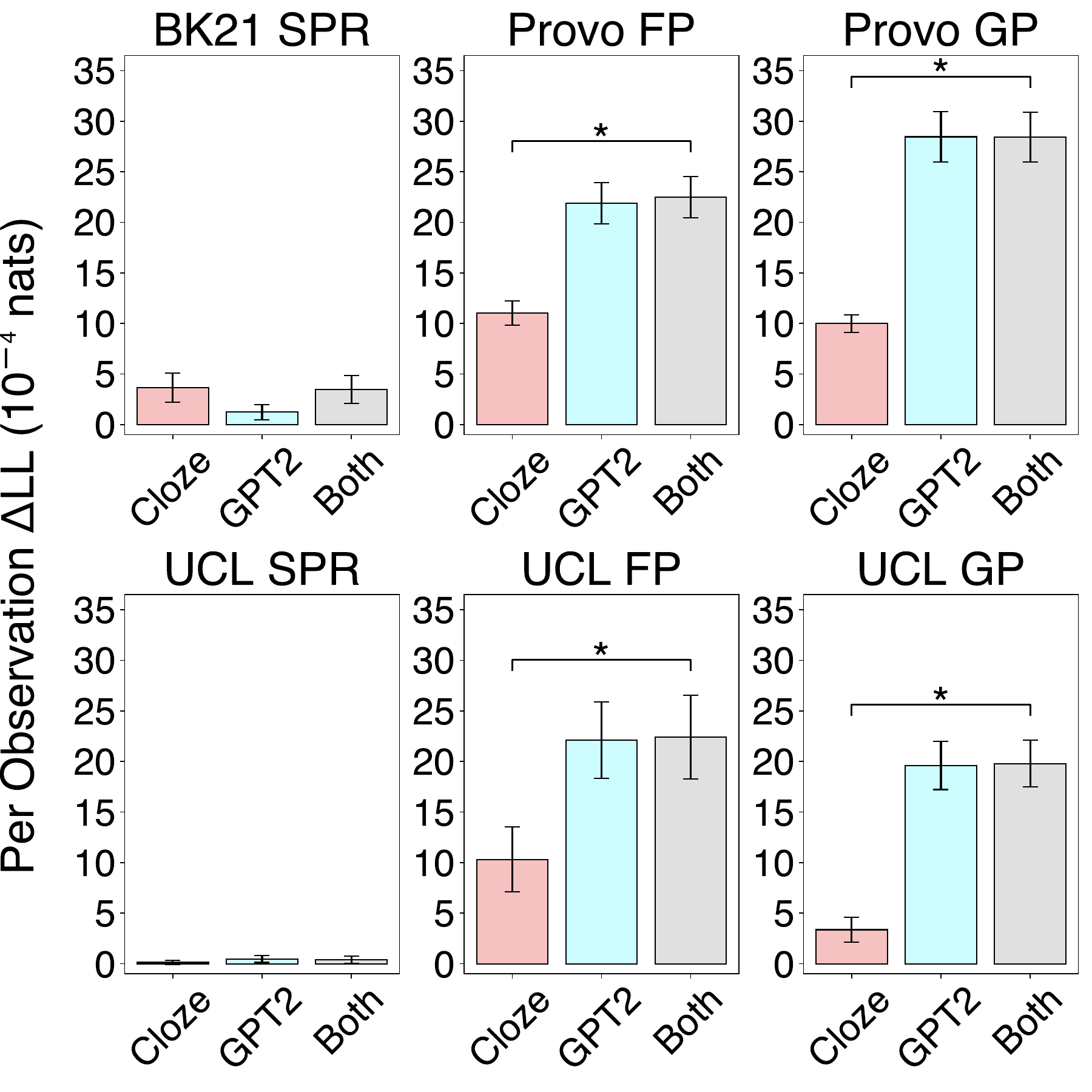}
    \caption{Increase in per-observation log likelihood over the baseline regression models due to including cloze surprisal, GPT2 surprisal, and both predictors, averaged over the 10 folds used in cross-validation. Error bars denote one standard error of the mean (SEM) across the 10 folds. Among the two comparisons of interest (Cloze vs.~Both; GPT2 vs.~Both), differences that achieve significance at the $0.05$ level by a paired permutation test under a 12-way Bonferroni correction (two comparisons on six measures) are marked with an asterisk.}
    \label{fig:exp1}
\end{figure}

\begin{table*}[ht!]
    \centering
    \begin{tabular}{S[table-format=4.0] S[table-format=3.1] S[table-format=3.1] S[table-format=3.1] S[table-format=3.1] S[table-format=3.1] S[table-format=3.1] }
    \toprule
    \multicolumn{1}{c}{$V$} & \multicolumn{1}{r}{$\prob{\wv_\tv}$} & \multicolumn{1}{r}{$\surp{\wv_\tv}$} & \multicolumn{1}{r}{$\surp{\wv_\tv}^{\frac{1}{2}}$} & \multicolumn{1}{r}{$\surp{\wv_\tv}^{\frac{3}{4}}$} & \multicolumn{1}{r}{$\surp{\wv_\tv}^{\frac{4}{3}}$} & \multicolumn{1}{r}{$\surp{\wv_\tv}^{2}$} \\ \midrule
    50 & 91.7 & 149.2 & 141.2 & 145.9 & 152.1 & \textbf{153.1} \\
    100 & 92.1 & 149.7 & 143.3 & 146.8 & 152.1 & \textbf{153.6} \\
    200 & 92.2 & 150.0 & 144.7 & 147.6 & 152.0 & \underline{\textbf{153.8}} \\
    500 & 92.4 & 150.1 & 146.1 & 148.3 & 152.0 & \textbf{153.7} \\
    1000 & 92.4 & 150.2 & 146.9 & 148.6 & 151.9 & \textbf{153.7} \\
    2000 & 92.6 & 150.2 & 147.3 & 149.0 & 151.6 & \textbf{153.4} \\ \bottomrule
    \end{tabular}
    \caption{The increase in regression model log likelihood due to including cloze probability calculated and aggregated over $\sim$50\% of the observations, as a function of different smoothing factors ($V$) and functional form to RT.}
    \label{tab:smoothing}
\end{table*}

\subsection{Results}
The results from cloze probabilities with different transforms in Table \ref{tab:smoothing} show that transforming probabilities into surprisal notably improves fit to RT; otherwise, the smoothing factor and the power transforms of surprisal did not have a very large effect.
Contrary to the findings of \citet{brothers_2021_word}, when other datasets are taken into consideration, the effect of cloze probability on reading time does not seem to be linear.
Throughout the remainder of this paper, we apply $\surp{\wv_\tv}^{2}$ with a smoothing factor of $V=200$ to all cloze probabilities, as this setup achieved the best fit to the six measures.\footnote{Prior work has also identified superlinear effects of LM surprisal \citep{meister_2021_revisiting, xu_2023_linearity, hoover2023plausibility}. It could be possible that a superlinear effect of cloze surprisal could be associated with production-related pressures, which have been proposed as an explanation for \citeauthor{meister_2021_revisiting}'s \citeyearpar{meister_2021_revisiting} results. However, the small differences between the linear and quadratic transformations of cloze surprisal are not enough to draw conclusions about the functional form of the relationship between surprisal and reading time.} %

Figure~\ref{fig:exp1} presents the main comparison between cloze and GPT2 surprisal.
On four out of six measures, GPT2 surprisal predicts RTs better than cloze surprisal, but not vice-versa, indicating that GPT2 surprisal subsumes the effect of cloze surprisal.
This corroborates earlier findings that show LM surprisal is a superior predictor of real-time comprehension data than cloze surprisal \citep{hofmann_2022_language, michaelov_2023_cloze, shain_2024_large}.
The two SPR measures that do not show a significant difference between the two predictors were not very well predicted by either cloze or GPT2 surprisal, which is possibly due to task-based differences between SPR and eyetracking.

\section{Experiment 2: Why Does LM Surprisal Predict RTs Better?}
Having established that GPT2 surprisal predicts RTs over and above cloze surprisal, the second experiment aims to identify where the modeling benefit of GPT2 surprisal comes from.
To this end, we manipulate GPT2 probabilities to test three hypotheses that may account for the difference between cloze responses and LM probabilities (Figure \ref{fig:schematic}).
The fit of manipulated GPT2 surprisal is compared against that of cloze surprisal following the same modeling setup as Experiment 1.
To the extent that each hypothesis is supported, we expect the fit to RT to decrease compared to what is reported in Figure \ref{fig:exp1}.

\subsection{Hypotheses and Manipulations}
\paragraph{Hypothesis 1 (H$_1$): Resolution} One practical limitation of conducting human studies to collect cloze responses is its cost: It can be prohibitively expensive to collect an ample number of responses over a text corpus.
Due to this limitation, cloze probabilities suffer from poor resolution, as they are based on counts from typically fewer than 100 responses, even if their underlying contextual expectations are higher-resolution.
In contrast, LMs can estimate probabilities for any arbitrary continuation using their vector representations, and therefore offer probabilities with much higher resolution.
To test the hypothesis that the difference in resolution accounts for the difference in the fit to RTs, we \textit{match} the resolution between cloze and GPT2 surprisal by basing GPT2 probabilities on the same number of samples as cloze responses.

More specifically, instead of directly calculating $\condprob{w_t}{w_{1..t-1}}$ from GPT2, we sample a set of words from the conditional distribution $\condprob{W}{w_{1..t-1}}$.
For each context, we sample the same number of words as the number of cloze responses ($N$) for each context.
Since LMs' distributions are typically defined over subword tokens rather than words, we iteratively sample $N$ token sequences from $\condprob{W}{w_{1..t-1}}$.
We first sample a token $t_0$ and its subsequent token $t_1$. If $t_1$ marks the end of a word (by containing a leading whitespace or being a punctuation) we take $t_0$ as a word sample.
Otherwise, we sample $t_2$ conditioned on $t_0$ and $t_1$ and check whether $t_2$ marks the end of a word.
We repeat this process up to $t_3$: If the end-of-word is not reached by $t_3$, we take the concatenation of $t_0$, $t_1$, and $t_2$ as our sample.
Given $N$ samples of words for each context, we calculate resolution-matched probabilities in a similar manner to Equation \ref{eqn:cloze}:

\begin{equation}
\label{eqn:h1}
    \mathsf{P}_{\text{H}_{1}}({\wv_\tv}\mid{\wv_{1...t-1}}) = \frac{C_{\wv_\tv}+1}{N+V},
\end{equation}
where $C_{\wv_\tv}$ is the number of times $w_t$ appears in the set of samples.
As with cloze probabilities, we apply $\surp{\wv_\tv}^{2}$ with a smoothing factor of $V=200$ prior to regression modeling.
To account for variance in the sampling process, we report median regression model performance over five runs.

Theoretically speaking, this manipulation aligns LM-based estimates with \citeauthor{smith_2011_cloze}'s \citeyearpar{smith_2011_cloze} view of the cloze task, where each response from a human subject is seen as a sample from their subjective probability distribution.\footnote{\citet{hao-etal-2020-probabilistic} claim neural LMs' probability estimates may be closer to humans' subjective probability distributions compared to $n$-gram probability estimates, based on their higher fit to RTs and correlation with cloze probabilities.}

\paragraph{Hypothesis 2 (H$_2$): Semantics} Another consequence of the LM's ability to consider any arbitrary continuation is that they can make extremely fine-grained distinctions between different words: They assign different probabilities to, for example, \textit{couch} and \textit{sofa}.
Human lexical predictions, however, are influenced by shared semantic features across possible alternatives \citep[among others]{federmeier_1999_rose, federmeier2022connecting, brothers_2023_multiple}.
Recent work making use of cloze responses that share semantic features has been able to predict processing difficulty more effectively than LM probabilities \citep{arkhipova2025meaning}.

In order to incorporate the influence of shared semantic features into GPT2's probabilities, we first cluster GPT2's vocabulary using $k$-means clustering with token embeddings.
Subsequently, we define the probability of $\wv_\tv$ as the probability of its cluster, which is the total probability of its members:
\begin{equation}
    \mathsf{P}_{\text{H}_{2}}({\wv_\tv}\mid{\wv_{1...t-1}}) = \sum_{w_i \in C_{\wv_\tv}} \condprob{w_i}{\wv_{1...t-1}},
\end{equation}
%
%
where $C_{\wv_\tv}$ denotes the cluster which $\wv_\tv$ belongs to.
Under this manipulation, if \textit{couch} and \textit{sofa} are assigned the same cluster, then they will both be assigned the same cluster probability, which is the sum of the probability of \textit{couch}, \textit{sofa}, and other words that belong to that cluster.\footnote{An anonymous reviewer points out that this implementation may be over-simplistic. An alternative could be to adjust probabilities such that words in larger clusters gain probability, while words in smaller clusters lose probability. However, LM probabilities would then make fine-grained distinctions between words, therefore making it inappropriate to test H$_2$.}
If $\wv_\tv$ consists of multiple subword tokens, we calculate its probability by taking the product of subword cluster probabilities.\footnote{The impact of subword tokenization is likely to be negligible, as very few words in the corpora were split by the GPT2 tokenizer (5.1\% for BK21, 5.9\% for Provo, and 1.9\% for UCL). See \citet{nair2023words} and \citet{giulianelli-etal-2024-proper} for more general discussions.}
We perform $k$-means clustering with $k \in \{20, 40, 80, 100, 500, 1000\}$, again reporting median regression model performance over five runs for each $k$, to account for the randomness in clustering. We validate our clustering approach and provide representative examples of cluster members in Appendix \ref{app:cluster_ex}.

This manipulation is also inspired by recent work that incorporates semantic proximity between alternative completions into LM predictability estimates \citep{meister-etal-2024-towards}, and clustering LM representations to operationalize semantic predictability \citep{jacobs2025uncovering}.

%
\paragraph{Hypothesis 3 (H$_3$): Frequency}
Another potential difference between cloze and LM surprisal is that human subjects may not ever produce low-frequency continuations to a context \citep{smith_2011_cloze}, while effects of word frequency during reading are well documented \citep[among others]{kliegl2004length, shain2024word}.
In contrast, LMs assign probabilities to arbitrary low-frequency continuations, which may account for the improved fit of GPT2 surprisal over cloze surprisal.
We test this hypothesis by constraining GPT2 to assign probabilities to only high-frequency tokens. 

First, we separate the subword vocabulary of GPT2 into two non-overlapping subsets $V_F$ (frequent) and $V_I$ (infrequent) depending on whether the token meets some frequency threshold.
Subsequently, all probability of tokens in $V_I$ are set to zero, and the probability of tokens in $V_F$ is re-normalized to sum to one.
Finally, a form of add-one smoothing is applied to allow log transform:
\begin{align}
    \mathsf{P}_{\text{H}_{3}}&({\wv_\tv}\mid{\wv_{1...t-1}}) = \notag \\
    & \begin{cases}
        \frac{\condprob{w_t}{\wv_{1...t-1}}}{\sum_{w_i \in V_F} \condprob{w_i}{\wv_{1...t-1}}} \times \frac{|V_F|}{|V_F|+1} & \text{if } \wv_\tv \in V_F \\
        \frac{1}{|V_F|+1} & \text{if } \wv_\tv \in V_I.
    \end{cases} \label{eqn:h3} 
\end{align}
If $w_t$ consists of multiple subword tokens, we calculate its probability by multiplying the probabilities of its subword tokens under Equation \ref{eqn:h3}. We test three different frequency thresholds based on \texttt{wordfreq} \citep{robyn_speer_2022_7199437}; $10^3$ occurrences per billion words, $10^4$ occurrences per billion words, and $10^5$ occurrences per billion words.

We note that while the three hypotheses are motivated by different aspects of the cloze task, they are not necessarily mutually exclusive.

\begin{figure}[t!]
    \centering
    \includegraphics[width=\linewidth]{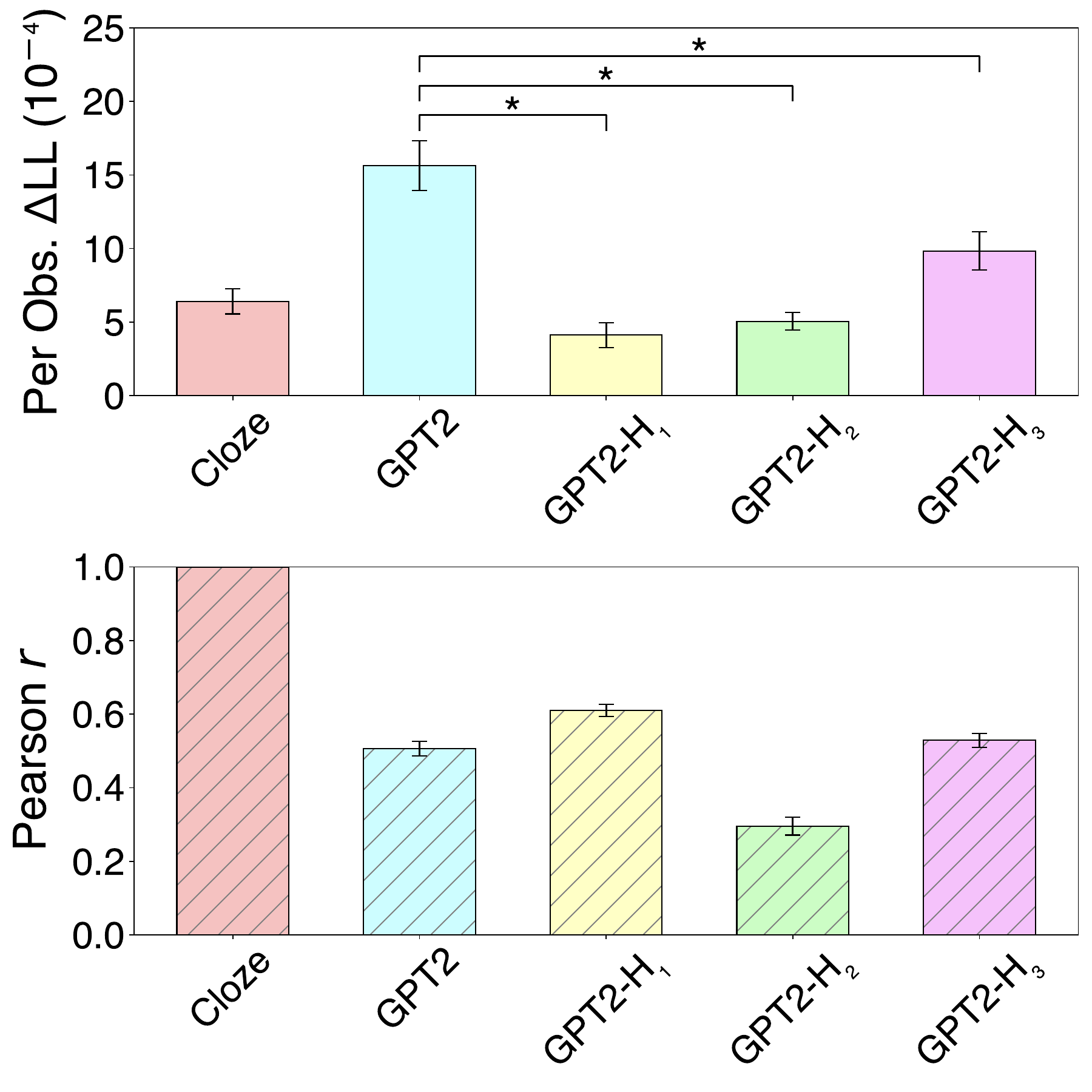}
    \caption{(\textbf{Top}) Increase in per-observation log likelihood over the baseline regression models due to including cloze surprisal, GPT2 surprisal, and manipulated variants of GPT2 surprisal, averaged over all 60 folds (10 folds of six measures) used in cross-validation. Error bars denote one SEM across all 60 folds. Among the three comparisons between GPT2 and its manipulated variants, differences that achieve significance at the $0.05$ level by a paired permutation test under a 3-way Bonferroni correction are marked with an asterisk. (\textbf{Bottom}) Pearson correlation between cloze probabilities and each set of GPT2-based probabilities, calculated over the three text corpora. Error bars denote 95\% confidence intervals derived by a permutation test.}
    \label{fig:exp2}
\end{figure}

\newlength{\oldtabcolsep}
\setlength{\oldtabcolsep}{\tabcolsep}
\setlength{\tabcolsep}{4pt}
\begin{figure*}[ht!]
    \centering
    \begin{tabular}{c|c|c}
       \textbf{H$_1$: Resolution} & \textbf{H$_2$: Semantics} & \textbf{H$_3$: Frequency} \\
       \includegraphics[width=0.315\linewidth]{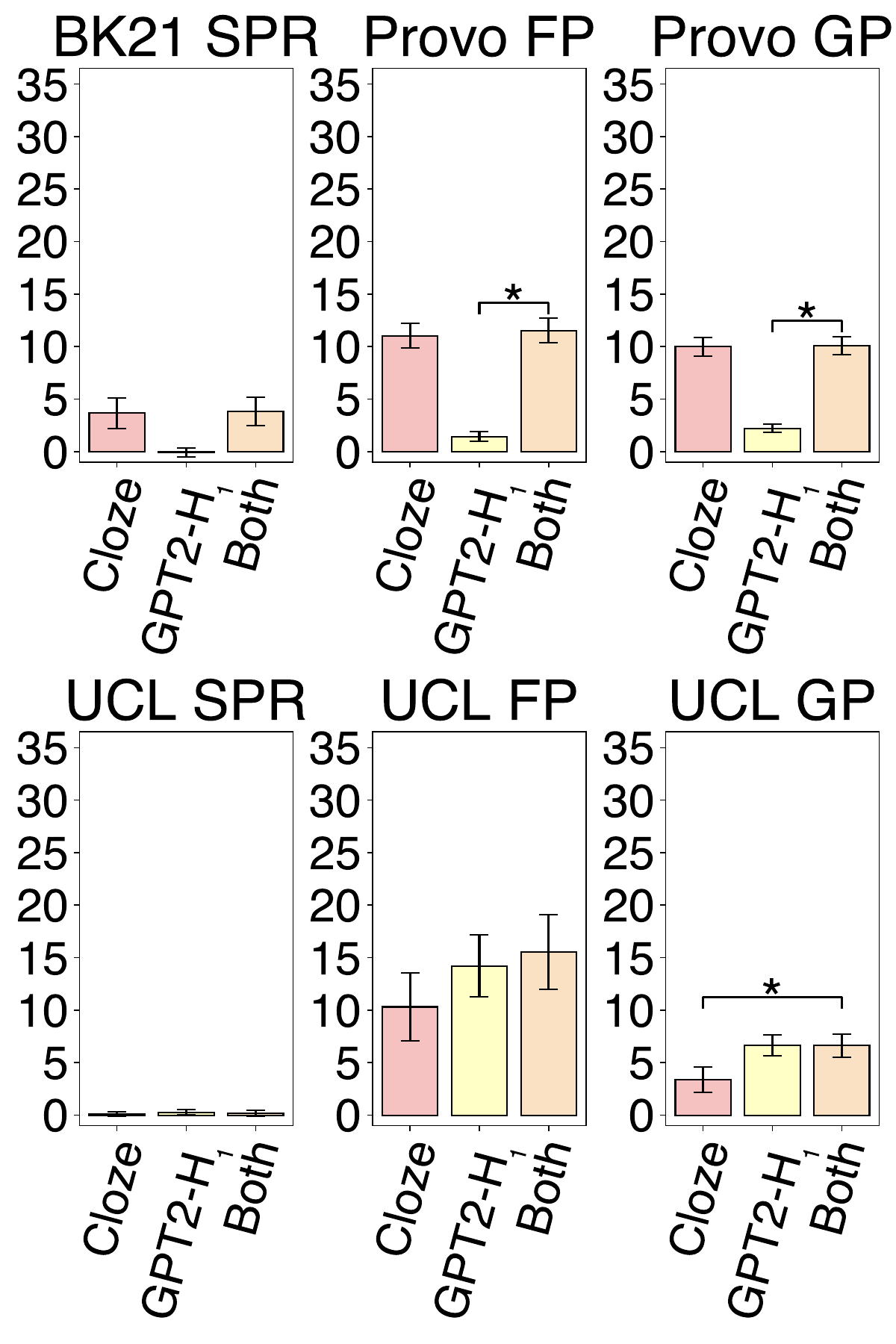} & \includegraphics[width=0.315\linewidth]{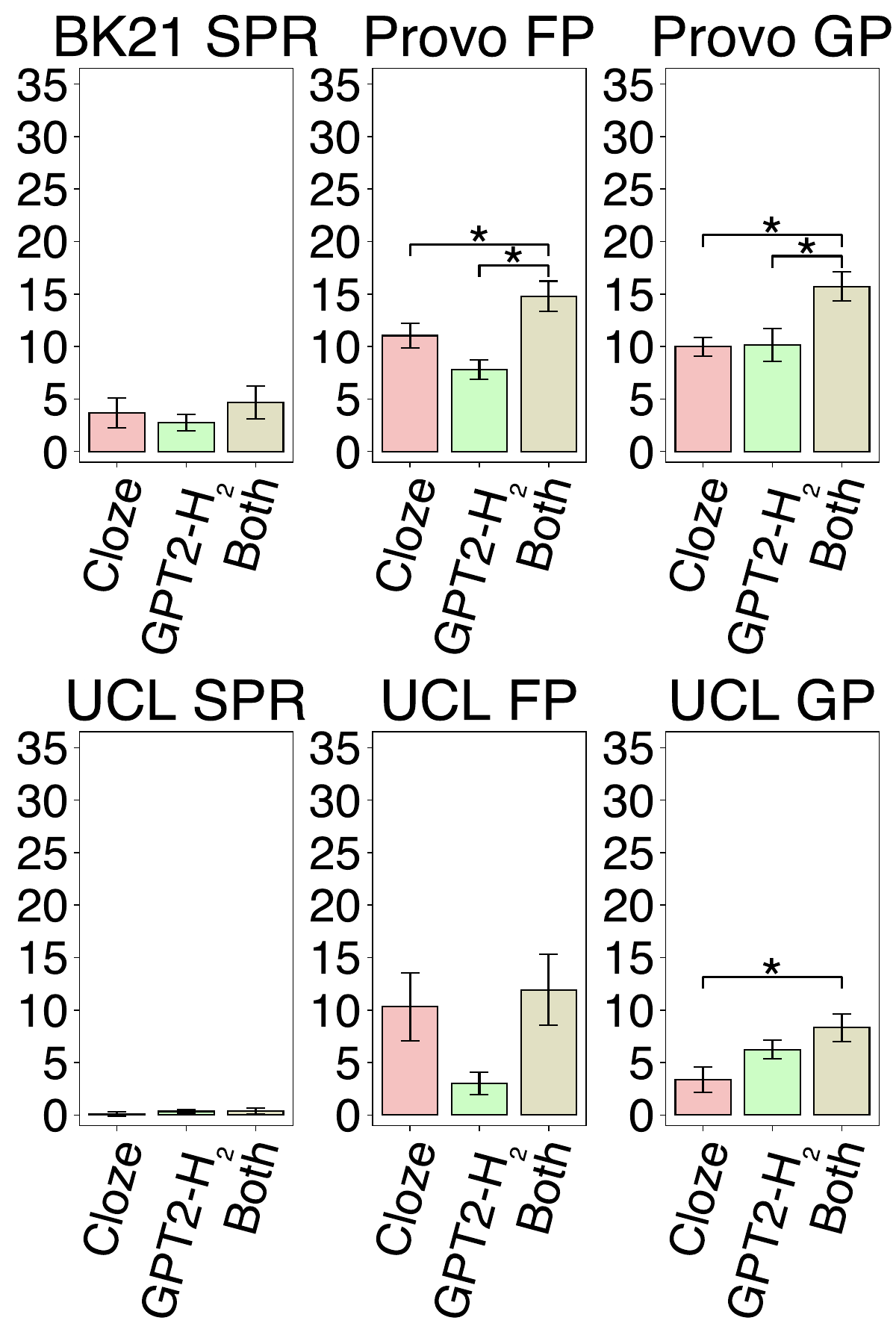} & \includegraphics[width=0.315\linewidth]{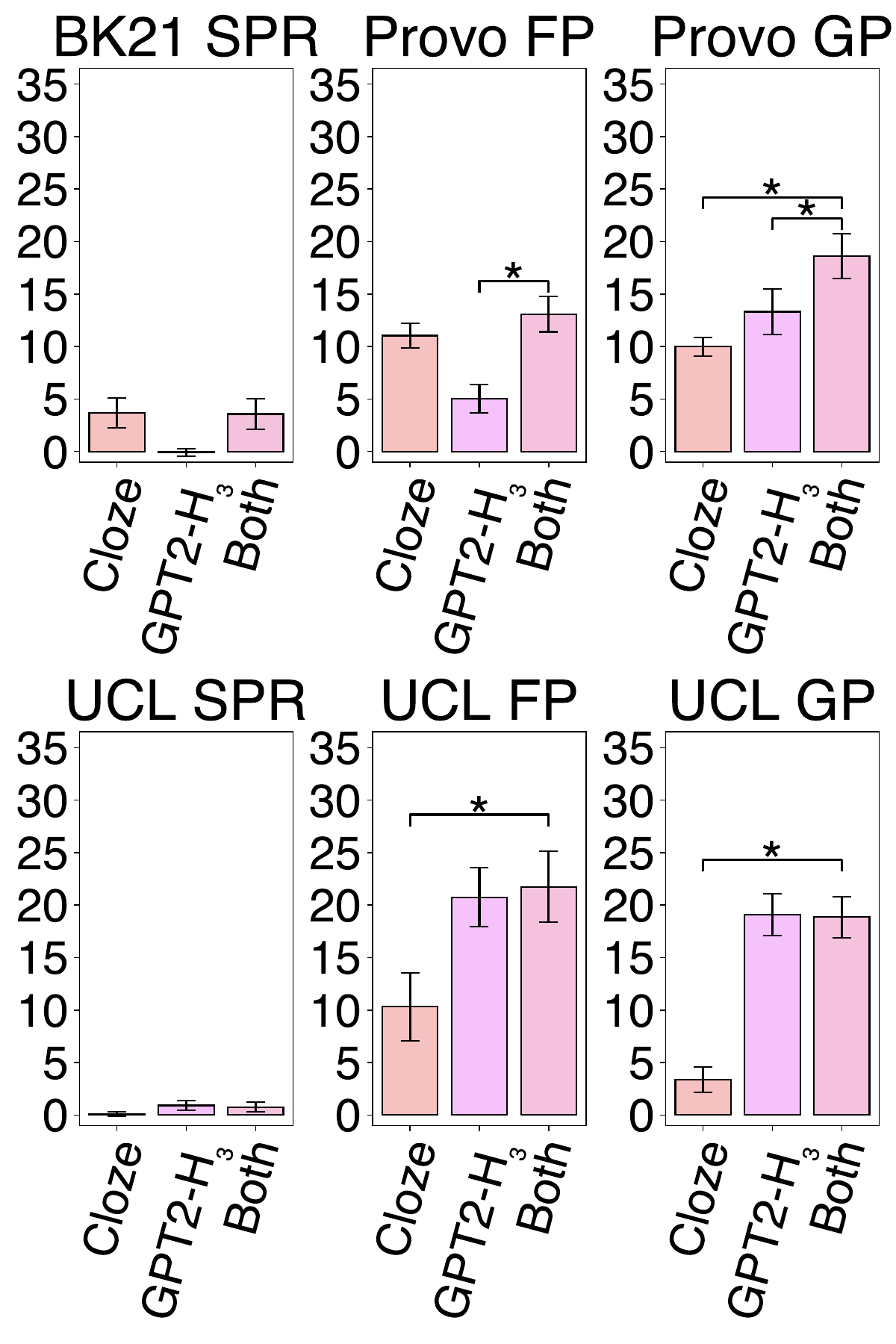}
    \end{tabular}
    \caption{Increase in per-observation log likelihood in $10^{-4}$ nats over the baseline regression models due to including cloze surprisal, manipulated GPT2 surprisal, and both predictors, averaged over the 10 folds used in cross-validation. Error bars denote one SEM across the 10 folds. Among the two comparisons of interest (Cloze vs.~Both; GPT2-H$_{\{1,2,3\}}$ vs.~Both), differences that achieve significance at the $0.05$ level by a paired permutation test under a 12-way Bonferroni correction (two comparisons on six measures) are marked with an asterisk.}
    \label{fig:exp2_bymeasure}
\end{figure*}

\subsection{Results}
\label{sec:e2_results}

The aggregate results in Figure \ref{fig:exp2} show that all three manipulations result in a significant decrease in fit to human RTs, providing support for all three hypotheses.\footnote{For H$_2$ and H$_3$, which required decisions about the number of clusters and the frequency threshold respectively, we report results using 80 clusters and a threshold of $10^4$ per billion as representative examples of the similar trend observed across settings. Results based on different numbers of clusters and frequency thresholds, alongside results from the five individual runs for H$_1$ and H$_2$, are reported in Appendix \ref{app:exp2_full}.}
In other words, GPT2 appears to yield stronger predictors of RTs than cloze responses due to its ability to assign highly fine-grained probabilities (H$_1$) and distinguish between semantically similar words (H$_2$) and low-frequency words (H$_3$).

The by-measure results in Figure \ref{fig:exp2_bymeasure} show that compared to results from unaltered GPT2 surprisal in Figure \ref{fig:exp1}, the decrease in fit to RT is the most apparent on the Provo ET measures, with cloze surprisal now explaining RTs over and above surprisal from the manipulated GPT2 variants.
We speculate that the effect is not as large on the UCL ET measures due to the different characteristics of the corpus.
The UCL corpus is a collection of short, isolated sentences that consist of all high-frequency words; therefore, most notably, our manipulation based on H$_{3}$ is unlikely to change the surprisal predictors by much.
\section{Experiment 3: Towards Combining Cloze Responses With LM Probabilities}
The previous experiment showed that GPT2's ability to assign fine-grained, `high-resolution' probabilities to the actual observed word $\wv_\tv$ provides an advantage for modeling human RTs.
In this experiment, we combine cloze responses with GPT2 probabilities through the link of similarity-adjusted surprisal \citep[SA surprisal;][c.f. cooccurrence-based smoothing; \citealp{essen1992cooccurrence}]{meister-etal-2024-towards}.
SA surprisal is based on the average probability of likely alternatives to the observed word $\wv_\tv$, which is weighted by their distance to $\wv_\tv$. Unlike H$_2$, SA surprisal considers attested alternative completions for the context, rather than words related in meaning to $\wv_\tv$.
It can therefore alleviate issues with the poor resolution of cloze responses; $\prob{\wv_\tv}$ can be calculated even if $\wv_\tv$ itself is not attested in the set of cloze responses.

By evaluating SA surprisal calculated based on the set of cloze responses (SA cloze surprisal), we ask two questions.
The first is whether SA cloze surprisal results in a different fit to RTs compared to `count-and-divide' cloze surprisal.
For example, SA cloze surprisal may capture the fact that a reader's reading may be facilitated by \textit{sofa}, even if they produced only \textit{couch} as their cloze response.
If so, this may suggest the need for a different method to derive surprisal from cloze responses.

The second comparison is against SA surprisal calculated based on the set of GPT2's responses (SA GPT2 surprisal).
By holding the underlying distance space and probabilities constant, this comparison asks whether the set of alternatives that GPT2 considers (i.e.~words that might be predicted instead of $\wv_{\tv}$) also provides a modeling benefit over the set of cloze responses.

\subsection{Methods}
SA surprisal is defined as:
\begin{equation}
    \mathsf{P}_{\text{S}}({\wv_\tv}\mid{\wv_{1...t-1}}) = \! \sum_{w' \in R_{w_{1..t-1}}} z(w_t, w')\condprob{w'}{w_{1...t-1}},
\end{equation}
where $z(\cdot)$ is a similarity function and $R_{w_{1..t-1}}$ is a multiset of responses given the context $w_{1..t-1}$.
For $z(\cdot)$, we used normalized cosine distance between token embeddings from GPT2.
For $R_{w_{1..t-1}}$, we used the set of cloze responses to calculate SA cloze surprisal, and the set of GPT2 samples matched in number (i.e.~samples collected to evaluate H$_{1}$) to calculate SA GPT2 surprisal.
For words that consist of multiple tokens, we mean-pooled across their embeddings to calculate their distance to $\wv_\tv$, following \citet{giulianelli-etal-2024-generalized}.
The conditional probability $\condprob{w'}{w_{1...t-1}}$ is calculated from GPT2.
Protocols for regression modeling and statistical significance testing follow those of the previous experiments.\footnote{We do not report results on the BK21 dataset as it does not provide the raw cloze responses to be used as $R_{w_{1..t-1}}$.}

\begin{figure}[t!]
    \centering
    \includegraphics[width=\linewidth]{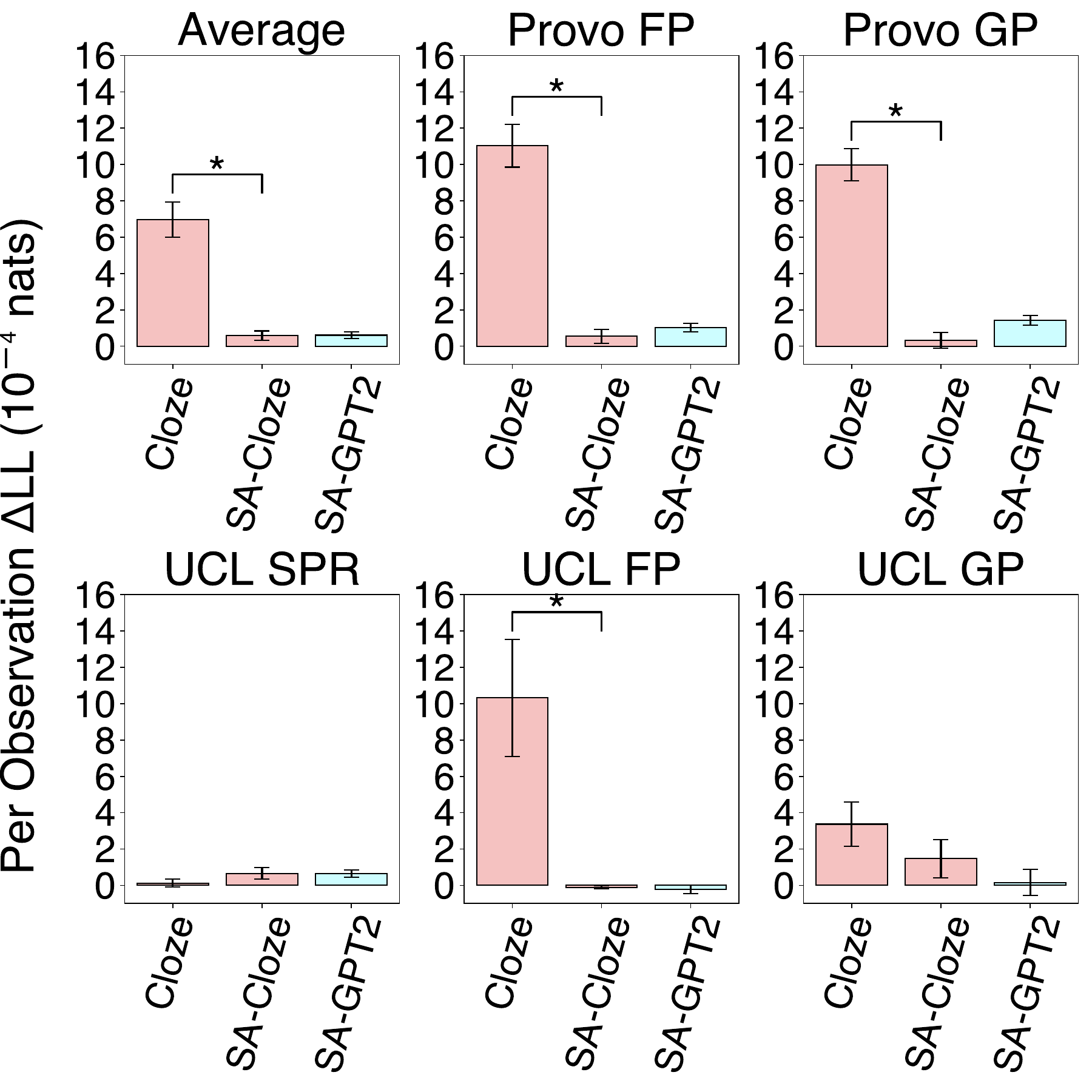}
    \caption{Increase in per-observation log likelihood, averaged over all folds used in cross-validation. Error bars denote one SEM across all folds. Differences that achieve significance at the $0.05$ level under a 10-way Bonferroni correction (two comparisons on five measures) are marked with an asterisk.}
    \label{fig:exp3}
\end{figure}

\subsection{Results}
Figure \ref{fig:exp3} shows that both versions of SA surprisal are poor predictors of RTs, indicating that count-and-divide is not an unreasonable method to convert cloze responses into probabilities.
Results are inconclusive about the quality of alternatives between cloze responses and GPT2 samples. 
We leave the exploration of methods for combining cloze responses with LM-based estimates to future work.

\section{Discussion and Conclusion}
To study how linguistic prediction shapes language comprehension, we need to quantify how predictable a word could be to a comprehender.
This work compares two very different methods for doing so against word-by-word reading times---using cloze responses and LM probabilities.
Consistently with previously reported findings, we first re-establish over multiple datasets that predictability estimates from LMs predict RTs better than those from cloze responses, at least for eyetracking data.\footnote{See \citet{buggypredictability} for more arguments about distinct cognitive processes recruited during self-paced reading and eyetracking and their consequences on predictability effects.}

However, instead of treating cloze responses as inferior purely on the basis of relatively poorer fit to RTs, we question where the modeling advantage of LM surprisal comes from.
To this end, we tested three hypotheses by manipulating GPT2 probabilities accordingly; namely that GPT2 does not suffer from poor resolution like the cloze task, and that its probabilities can distinguish between semantically similar words and low-frequency words.
Our results support all three hypotheses.

Perhaps the most noteworthy finding is that when LMs are used to generate cloze-like estimates from an underlying distribution (i.e.~the resolution is lowered), GPT2 surprisal is no longer a stronger predictor of RTs than cloze surprisal.
This suggests that cloze surprisal that is based on more responses than are typically collected may be a stronger predictor of RTs than the cloze surprisal evaluated in this work.
Nonetheless, simply collecting more responses will not address the inherent limitations of the cloze task, which stem from the fact that it is an untimed production task.
%
As such, we call for alternatives to the cloze task, such that it can better capture predictions during real-time language processing.
For example, in addition to the traditional cloze task, timed versions of the cloze task \citep{staub_2015_influence} or a maze-like variant may help control for factors like conscious reflection that can influence the responses.

At the same time, it remains to be seen whether human-like prediction is as sensitive to the fine-grained distinctions made by LM probabilities.
Future experiments should target whether humans' expectations differentiate between e.g.~semantically related words or low-frequency words---perhaps using stimuli informed by LM probabilities.
This will help ascertain whether LM surprisal indeed predicts RTs because it captures human-like aspects of prediction.
While neural network-based LMs offer a lot of convenience for applications in cognitive modeling, what drives their next-word predictions is not fully understood. 
Therefore, there remains a possibility that their probabilities capture human data during real-time processing due to reasons unrelated to human-like prediction.

From a theoretical perspective, it is also important to understand the empirical limitations of prediction in language comprehension \citep[e.g.][]{huang_2024_large, timkey_2025_eye, staub2025predictability}.
Therefore, instead of using LM probabilities as a potential one-size-fits-all explanation of all measures of processing effort during language comprehension, we encourage future modeling approaches to be more explicit about the links between various stages of probabilistic inference and individual measures.
An example of such class of models is predictive coding models, which not only implement a neurobiologically plausible approach to inference \citep{clark_2013_whatever}, but also offer cognitively interpretable measures that can be linked to different aspects of processing behavior when applied to language comprehension \citep{nour_eddine_predictive_2024, ohams2026predictive}.

%


%
\section*{Limitations}
This work relied on language models trained on English text and data from human subjects that are native speakers of English.
Therefore, it remains to be seen whether the findings will generalize to other language models and data collected in other languages.
Although multilingual language models and reading time datasets exist, we are not aware of any cross-linguistic datasets of cloze responses aligned to reading time data. 
In our manipulations for Hypotheses 2 and 3, we relied on multiplying token-level probabilities, instead of applying corrections used for unaltered word probabilities, since these corrections are not be likely to introduce major qualitative changes to our results \citep{oh_2024_leading, pimentel_2024_compute}.
%

\section*{Ethical Considerations}
This work used reading time data collected as part of previously published research.
We refer readers to the respective publications for the data collection and validation procedures.
This work used these datasets for their intended purpose---to study human sentence processing---and therefore we foresee no potential risks associated with this work.

\section*{Acknowledgments}

We thank Philip Resnik, Samer Nour Eddine, Colin Phillips, Cassandra Jacobs, members of the UMD psycholinguistics community, as well as the audience at the 39th Annual Conference on Human Sentence Processing for valuable discussion, and the ARR reviewers for constructive feedback.
This material is based on work supported by the NSF GRFP (No.~DGE 2236417) to Sathvik Nair.
Research for this paper was made possible by a Start-Up Grant (No.~03INS002748C420) from Nanyang Technological University, Singapore to Byung-Doh Oh.


\bibliography{custom}

\begin{thebibliography}{59}
\providecommand{\natexlab}[1]{#1}

\bibitem[{Arkhipova et~al.(2025)Arkhipova, Lopopolo, Vasishth, and Rabovsky}]{arkhipova2025meaning}
Yana Arkhipova, Alessandro Lopopolo, Shravan Vasishth, and Milena Rabovsky. 2025.
\newblock \href {https://doi.org/10.1101/2025.04.29.651301} {When meaning matters most: Rethinking cloze probability in {N400} research}.
\newblock \emph{bioRxiv}.

\bibitem[{Bates et~al.(2015)Bates, M{\"{a}}chler, Bolker, and Walker}]{bates_2015_fitting}
Douglas Bates, Martin M{\"{a}}chler, Ben Bolker, and Steve Walker. 2015.
\newblock \href {https://doi.org/10.18637/jss.v067.i01} {{Fitting linear mixed-effects models using lme4}}.
\newblock \emph{Journal of Statistical Software}, 67(1):1--48.

\bibitem[{Biderman et~al.(2023)Biderman, Schoelkopf, Anthony, Bradley, O'Brien, Hallahan, Khan, Purohit, Prashanth, Raff, Skowron, Sutawika, and van~der Wal}]{biderman_2023_pythia}
Stella Biderman, Hailey Schoelkopf, Quentin~Gregory Anthony, Herbie Bradley, Kyle O'Brien, Eric Hallahan, Mohammad~Aflah Khan, Shivanshu Purohit, USVSN~Sai Prashanth, Edward Raff, Aviya Skowron, Lintang Sutawika, and Oskar van~der Wal. 2023.
\newblock \href {https://proceedings.mlr.press/v202/biderman23a.html} {Pythia: A suite for analyzing large language models across training and scaling}.
\newblock In \emph{Proceedings of the 40th International Conference on Machine Learning}, volume 202, pages 2397--2430.

\bibitem[{Brothers and Kuperberg(2021)}]{brothers_2021_word}
Trevor Brothers and Gina~R. Kuperberg. 2021.
\newblock \href {https://doi.org/10.1016/j.jml.2020.104174} {{Word predictability effects are linear, not logarithmic: Implications for probabilistic models of sentence comprehension}}.
\newblock \emph{Journal of Memory and Language}, 116:104174.

\bibitem[{Brothers et~al.(2023)Brothers, Morgan, Yacovone, and Kuperberg}]{brothers_2023_multiple}
Trevor Brothers, Emily Morgan, Anthony Yacovone, and Gina Kuperberg. 2023.
\newblock \href {https://doi.org/10.1016/j.cognition.2023.105602} {Multiple predictions during language comprehension: Friends, foes, or indifferent companions?}
\newblock \emph{Cognition}, 241:105602.

\bibitem[{Buggy et~al.(2026)Buggy, Cho, Shain, and Staub}]{buggypredictability}
Ryan Buggy, Stephanie Cho, Cory Shain, and Adrian Staub. 2026.
\newblock \href {https://doi.org/10.31234/osf.io/kt3sw_v1} {Predictability effects in natural reading are logarithmic, not linear: Evidence from an eye-movement replication of {Brothers and Kuperberg} (2021)}.
\newblock \emph{PsyArXiv}, Kt3sw\_v1.

\bibitem[{Clark(2013)}]{clark_2013_whatever}
Andy Clark. 2013.
\newblock \href {https://doi.org/10.1017/S0140525X12000477} {Whatever next? {P}redictive brains, situated agents, and the future of cognitive science}.
\newblock \emph{Behavioral and Brain Sciences}, 36(3):181--204.

\bibitem[{de~Varda et~al.(2024)de~Varda, Marelli, and Amenta}]{devarda_2024_cloze}
Andrea~Gregor de~Varda, Marco Marelli, and Simona Amenta. 2024.
\newblock \href {https://doi.org/10.3758/s13428-023-02261-8} {{Cloze probability, predictability ratings, and computational estimates for 205 English sentences, aligned with existing EEG and reading time data}}.
\newblock \emph{Behavior Research Methods}, 56:5190--5213.

\bibitem[{Ehrlich and Rayner(1981)}]{ehrlich_1981_contextual}
Susan~F. Ehrlich and Keith Rayner. 1981.
\newblock \href {https://doi.org/10.1016/S0022-5371(81)90220-6} {{Contextual effects on word perception and eye movements during reading}}.
\newblock \emph{Journal of Verbal Learning and Verbal Behavior}, 20(6):641--655.

\bibitem[{Essen and Steinbiss(1992)}]{essen1992cooccurrence}
Ute Essen and Volker Steinbiss. 1992.
\newblock \href {https://doi.org/10.1109/ICASSP.1992.225947} {Cooccurrence smoothing for stochastic language modeling}.
\newblock In \emph{Proceedings of IEEE International Conference on Acoustics, Speech, and Signal Processing}, volume~1, pages 161--164.

\bibitem[{Federmeier(2022)}]{federmeier2022connecting}
Kara~D. Federmeier. 2022.
\newblock \href {https://doi.org/10.1111/psyp.13940} {Connecting and considering: Electrophysiology provides insights into comprehension}.
\newblock \emph{Psychophysiology}, 59(1):e13940.

\bibitem[{Federmeier and Kutas(1999)}]{federmeier_1999_rose}
Kara~D. Federmeier and Marta Kutas. 1999.
\newblock \href {https://doi.org/10.1006/jmla.1999.2660} {A rose by any other name: Long-term memory structure and sentence processing}.
\newblock \emph{Journal of Memory and Language}, 41(4):469--495.

\bibitem[{Frank et~al.(2013)Frank, Fernandez~Monsalve, Thompson, and Vigliocco}]{frank_2013_reading}
Stefan~L. Frank, Irene Fernandez~Monsalve, Robin~L. Thompson, and Gabriella Vigliocco. 2013.
\newblock \href {https://doi.org/10.3758/s13428-012-0313-y} {{Reading time data for evaluating broad-coverage models of English sentence processing}}.
\newblock \emph{Behavior Research Methods}, 45(4):1182--1190.

\bibitem[{Giulianelli et~al.(2024{\natexlab{a}})Giulianelli, Malagutti, Gastaldi, DuSell, Vieira, and Cotterell}]{giulianelli-etal-2024-proper}
Mario Giulianelli, Luca Malagutti, Juan~Luis Gastaldi, Brian DuSell, Tim Vieira, and Ryan Cotterell. 2024{\natexlab{a}}.
\newblock \href {https://doi.org/10.18653/v1/2024.emnlp-main.1032} {On the proper treatment of tokenization in psycholinguistics}.
\newblock In \emph{Proceedings of the 2024 Conference on Empirical Methods in Natural Language Processing}, pages 18556--18572.

\bibitem[{Giulianelli et~al.(2024{\natexlab{b}})Giulianelli, Opedal, and Cotterell}]{giulianelli-etal-2024-generalized}
Mario Giulianelli, Andreas Opedal, and Ryan Cotterell. 2024{\natexlab{b}}.
\newblock \href {https://doi.org/10.18653/v1/2024.findings-emnlp.682} {Generalized measures of anticipation and responsivity in online language processing}.
\newblock In \emph{Findings of the Association for Computational Linguistics: EMNLP 2024}, pages 11648--11669.

\bibitem[{Giulianelli et~al.(2026)Giulianelli, Wallbridge, Cotterell, and Fernández}]{giulianelli_2026_incremental}
Mario Giulianelli, Sarenne Wallbridge, Ryan Cotterell, and Raquel Fernández. 2026.
\newblock \href {https://doi.org/10.1016/j.jml.2025.104715} {Incremental alternative sampling as a lens into the temporal and representational resolution of linguistic prediction}.
\newblock \emph{Journal of Memory and Language}, 148:104715.

\bibitem[{Gokaslan and Cohen(2019)}]{gokaslan_2019_openwebtext}
Aaron Gokaslan and Vanya Cohen. 2019.
\newblock \href {http://Skylion007.github.io/OpenWebTextCorpus} {{OpenWebText Corpus}}.

\bibitem[{Grattafiori et~al.(2024)Grattafiori, Dubey, Jauhri, Pandey, Kadian, Al-Dahle, Letman, Mathur, Schelten, Vaughan, Yang, Fan, Goyal, Hartshorn, Yang, Mitra, Sravankumar, Korenev, Hinsvark, Rao, Zhang, Rodriguez, Gregerson, Spataru, Roziere, Biron, Tang, Chern, Caucheteux, Nayak, Bi, Marra, McConnell, Keller, Touret, Wu, Wong, Ferrer, Nikolaidis, Allonsius, Song, Pintz, Livshits, Wyatt, Esiobu, Choudhary, Mahajan, Garcia-Olano, Perino, Hupkes, Lakomkin, AlBadawy, Lobanova, Dinan, Smith, Radenovic, Guzmán, Zhang, Synnaeve, Lee, Anderson, Thattai, Nail, Mialon, Pang, Cucurell, Nguyen, Korevaar, Xu, Touvron, Zarov, Ibarra, Kloumann, Misra, Evtimov, Zhang, Copet, Lee, Geffert, Vranes, Park, Mahadeokar, Shah, van~der Linde, Billock, Hong, Lee, Fu, Chi, Huang, Liu, Wang, Yu, Bitton, Spisak, Park, Rocca, Johnstun, Saxe, Jia, Alwala, Prasad, Upasani, Plawiak, Li, Heafield, Stone, El-Arini, Iyer, Malik, Chiu, Bhalla, Lakhotia, Rantala-Yeary, van~der Maaten, Chen, Tan, Jenkins, Martin, Madaan, Malo, Blecher,
  Landzaat, de~Oliveira, Muzzi, Pasupuleti, Singh, Paluri, Kardas, Tsimpoukelli, Oldham, Rita, Pavlova, Kambadur, Lewis, Si, Singh, Hassan, Goyal, Torabi, Bashlykov, Bogoychev, Chatterji, Zhang, Duchenne, Çelebi, Alrassy, Zhang, Li, Vasic, Weng, Bhargava, Dubal, Krishnan, Koura, Xu, He, Dong, Srinivasan, Ganapathy, Calderer, Cabral, Stojnic, Raileanu, Maheswari, Girdhar, Patel, Sauvestre, Polidoro, Sumbaly, Taylor, Silva, Hou, Wang, Hosseini, Chennabasappa, Singh, Bell, Kim, Edunov, Nie, Narang, Raparthy, Shen, Wan, Bhosale, Zhang, Vandenhende, Batra, Whitman, Sootla, Collot, Gururangan, Borodinsky, Herman, Fowler, Sheasha, Georgiou, Scialom, Speckbacher, Mihaylov, Xiao, Karn, Goswami, Gupta, Ramanathan, Kerkez, Gonguet, Do, Vogeti, Albiero, Petrovic, Chu, Xiong, Fu, Meers, Martinet, Wang, Wang, Tan, Xia, Xie, Jia, Wang, Goldschlag, Gaur, Babaei, Wen, Song, Zhang, Li, Mao, Coudert, Yan, Chen, Papakipos, Singh, Srivastava, Jain, Kelsey, Shajnfeld, Gangidi, Victoria, Goldstand, Menon, Sharma, Boesenberg,
  Baevski, Feinstein, Kallet, Sangani, Teo, Yunus, Lupu, Alvarado, Caples, Gu, Ho, Poulton, Ryan, Ramchandani, Dong, Franco, Goyal, Saraf, Chowdhury, Gabriel, Bharambe, Eisenman, Yazdan, James, Maurer, Leonhardi, Huang, Loyd, Paola, Paranjape, Liu, Wu, Ni, Hancock, Wasti, Spence, Stojkovic, Gamido, Montalvo, Parker, Burton, Mejia, Liu, Wang, Kim, Zhou, Hu, Chu, Cai, Tindal, Feichtenhofer, Gao, Civin, Beaty, Kreymer, Li, Adkins, Xu, Testuggine, David, Parikh, Liskovich, Foss, Wang, Le, Holland, Dowling, Jamil, Montgomery, Presani, Hahn, Wood, Le, Brinkman, Arcaute, Dunbar, Smothers, Sun, Kreuk, Tian, Kokkinos, Ozgenel, Caggioni, Kanayet, Seide, Florez, Schwarz, Badeer, Swee, Halpern, Herman, Sizov, Guangyi, Zhang, Lakshminarayanan, Inan, Shojanazeri, Zou, Wang, Zha, Habeeb, Rudolph, Suk, Aspegren, Goldman, Zhan, Damlaj, Molybog, Tufanov, Leontiadis, Veliche, Gat, Weissman, Geboski, Kohli, Lam, Asher, Gaya, Marcus, Tang, Chan, Zhen, Reizenstein, Teboul, Zhong, Jin, Yang, Cummings, Carvill, Shepard, McPhie,
  Torres, Ginsburg, Wang, Wu, U, Saxena, Khandelwal, Zand, Matosich, Veeraraghavan, Michelena, Li, Jagadeesh, Huang, Chawla, Huang, Chen, Garg, A, Silva, Bell, Zhang, Guo, Yu, Moshkovich, Wehrstedt, Khabsa, Avalani, Bhatt, Mankus, Hasson, Lennie, Reso, Groshev, Naumov, Lathi, Keneally, Liu, Seltzer, Valko, Restrepo, Patel, Vyatskov, Samvelyan, Clark, Macey, Wang, Hermoso, Metanat, Rastegari, Bansal, Santhanam, Parks, White, Bawa, Singhal, Egebo, Usunier, Mehta, Laptev, Dong, Cheng, Chernoguz, Hart, Salpekar, Kalinli, Kent, Parekh, Saab, Balaji, Rittner, Bontrager, Roux, Dollar, Zvyagina, Ratanchandani, Yuvraj, Liang, Alao, Rodriguez, Ayub, Murthy, Nayani, Mitra, Parthasarathy, Li, Hogan, Battey, Wang, Howes, Rinott, Mehta, Siby, Bondu, Datta, Chugh, Hunt, Dhillon, Sidorov, Pan, Mahajan, Verma, Yamamoto, Ramaswamy, Lindsay, Lindsay, Feng, Lin, Zha, Patil, Shankar, Zhang, Zhang, Wang, Agarwal, Sajuyigbe, Chintala, Max, Chen, Kehoe, Satterfield, Govindaprasad, Gupta, Deng, Cho, Virk, Subramanian, Choudhury,
  Goldman, Remez, Glaser, Best, Koehler, Robinson, Li, Zhang, Matthews, Chou, Shaked, Vontimitta, Ajayi, Montanez, Mohan, Kumar, Mangla, Ionescu, Poenaru, Mihailescu, Ivanov, Li, Wang, Jiang, Bouaziz, Constable, Tang, Wu, Wang, Wu, Gao, Kleinman, Chen, Hu, Jia, Qi, Li, Zhang, Zhang, Adi, Nam, Yu, Wang, Zhao, Hao, Qian, Li, He, Rait, DeVito, Rosnbrick, Wen, Yang, Zhao, and Ma}]{grattafiori_2024_llama}
Aaron Grattafiori, Abhimanyu Dubey, Abhinav Jauhri, Abhinav Pandey, Abhishek Kadian, Ahmad Al-Dahle, Aiesha Letman, Akhil Mathur, Alan Schelten, Alex Vaughan, Amy Yang, Angela Fan, Anirudh Goyal, Anthony Hartshorn, Aobo Yang, Archi Mitra, Archie Sravankumar, Artem Korenev, Arthur Hinsvark, and 542 others. 2024.
\newblock \href {https://arxiv.org/abs/2407.21783} {{The Llama 3 herd of models}}.
\newblock \emph{arXiv preprint}, arXiv:2407.21783v2.

\bibitem[{Hao et~al.(2020)Hao, Mendelsohn, Sterneck, Martinez, and Frank}]{hao-etal-2020-probabilistic}
Yiding Hao, Simon Mendelsohn, Rachel Sterneck, Randi Martinez, and Robert Frank. 2020.
\newblock \href {https://doi.org/10.18653/v1/2020.cmcl-1.10} {Probabilistic predictions of people perusing: Evaluating metrics of language model performance for psycholinguistic modeling}.
\newblock In \emph{Proceedings of the Workshop on Cognitive Modeling and Computational Linguistics}, pages 75--86.

\bibitem[{Heafield et~al.(2013)Heafield, Pouzyrevsky, Clark, and Koehn}]{heafield_2013_scalable}
Kenneth Heafield, Ivan Pouzyrevsky, Jonathan~H. Clark, and Philipp Koehn. 2013.
\newblock \href {https://www.aclweb.org/anthology/P13-2121/} {{Scalable modified Kneser-Ney language model estimation}}.
\newblock In \emph{Proceedings of the 51st Annual Meeting of the Association for Computational Linguistics}, pages 690--696.

\bibitem[{Hofmann et~al.(2022)Hofmann, Remus, Biemann, Radach, and Kuchinke}]{hofmann_2022_language}
Markus~J. Hofmann, Steffen Remus, Chris Biemann, Ralph Radach, and Lars Kuchinke. 2022.
\newblock \href {https://doi.org/10.3389/frai.2021.730570} {Language models explain word reading times better than empirical predictability}.
\newblock \emph{Frontiers in Artificial Intelligence}, 4:730570.

\bibitem[{Hoover et~al.(2023)Hoover, Sonderegger, Piantadosi, and O'Donnell}]{hoover2023plausibility}
Jacob~Louis Hoover, Morgan Sonderegger, Steven~T. Piantadosi, and Timothy~J. O'Donnell. 2023.
\newblock \href {https://doi.org/10.1162/opmi_a_00086} {The plausibility of sampling as an algorithmic theory of sentence processing}.
\newblock \emph{Open Mind}, 7:350--391.

\bibitem[{Huang et~al.(2024)Huang, Arehalli, Kugemoto, Muxica, Prasad, Dillon, and Linzen}]{huang_2024_large}
Kuan-Jung Huang, Suhas Arehalli, Mari Kugemoto, Christian Muxica, Grusha Prasad, Brian Dillon, and Tal Linzen. 2024.
\newblock \href {https://doi.org/10.1016/j.jml.2024.104510} {Large-scale benchmark yields no evidence that language model surprisal explains syntactic disambiguation difficulty}.
\newblock \emph{Journal of Memory and Language}, 137:104510.

\bibitem[{Hubert and Arabie(1985)}]{hubert1985comparing}
Lawrence Hubert and Phipps Arabie. 1985.
\newblock \href {https://doi.org/10.1007/BF01908075} {Comparing partitions}.
\newblock \emph{Journal of Classification}, 2:193--218.

\bibitem[{Jacobs et~al.(2024)Jacobs, Grobol, and Tsang}]{jacobs_2024_large}
Cassandra~L. Jacobs, Loïc Grobol, and Alvin Tsang. 2024.
\newblock \href {https://arxiv.org/abs/2410.12057} {Large-scale cloze evaluation reveals that token prediction tasks are neither lexically nor semantically aligned}.
\newblock \emph{arXiv preprint}, arXiv:2410.12057v2.

\bibitem[{Jacobs et~al.(2025)Jacobs, Hubbard, Grobol, and Federmeier}]{jacobs2025uncovering}
Cassandra~L. Jacobs, Ryan~J. Hubbard, Lo{\"\i}c Grobol, and Kara~D. Federmeier. 2025.
\newblock \href {https://doi.org/10.1016/j.jml.2025.104653} {Uncovering patterns of semantic predictability in sentence processing}.
\newblock \emph{Journal of Memory and Language}, 144:104653.

\bibitem[{Kliegl et~al.(2004)Kliegl, Grabner, Rolfs, and Engbert}]{kliegl2004length}
Reinhold Kliegl, Ellen Grabner, Martin Rolfs, and Ralf Engbert. 2004.
\newblock \href {https://doi.org/10.1080/09541440340000213} {Length, frequency, and predictability effects of words on eye movements in reading}.
\newblock \emph{European Journal of Cognitive Psychology}, 16(1-2):262--284.

\bibitem[{Kuribayashi et~al.(2025)Kuribayashi, Oseki, Taieb, Inui, and Baldwin}]{kuribayashi_2025_large}
Tatsuki Kuribayashi, Yohei Oseki, Souhaib~Ben Taieb, Kentaro Inui, and Timothy Baldwin. 2025.
\newblock \href {https://doi.org/10.1162/TACL.a.58} {Large language models are human-like internally}.
\newblock \emph{Transactions of the Association for Computational Linguistics}, 13:1743--1766.

\bibitem[{Kutas and Hillyard(1984)}]{kutas_1984_brain}
Marta Kutas and Steven~A. Hillyard. 1984.
\newblock \href {https://doi.org/10.1038/307161a0} {{Brain potentials during reading reflect word expectancy and semantic association}}.
\newblock \emph{Nature}, 307(5947):161--163.

\bibitem[{Luke and Christianson(2018)}]{luke_2018_provo}
Steven~G. Luke and Kiel Christianson. 2018.
\newblock \href {https://doi.org/10.3758/s13428-017-0908-4} {{The Provo Corpus: A large eye-tracking corpus with predictability norms}}.
\newblock \emph{Behavior Research Methods}, 50(2):826--833.

\bibitem[{Meister et~al.(2024)Meister, Giulianelli, and Pimentel}]{meister-etal-2024-towards}
Clara Meister, Mario Giulianelli, and Tiago Pimentel. 2024.
\newblock \href {https://doi.org/10.18653/v1/2024.emnlp-main.921} {Towards a similarity-adjusted surprisal theory}.
\newblock In \emph{Proceedings of the 2024 Conference on Empirical Methods in Natural Language Processing}, pages 16485--16498.

\bibitem[{Meister et~al.(2021)Meister, Pimentel, Haller, J{\"a}ger, Cotterell, and Levy}]{meister_2021_revisiting}
Clara Meister, Tiago Pimentel, Patrick Haller, Lena J{\"a}ger, Ryan Cotterell, and Roger Levy. 2021.
\newblock \href {https://doi.org/10.18653/v1/2021.emnlp-main.74} {Revisiting the {U}niform {I}nformation {D}ensity hypothesis}.
\newblock In \emph{Proceedings of the 2021 Conference on Empirical Methods in Natural Language Processing}, pages 963--980.

\bibitem[{Michaelov et~al.(2023)Michaelov, Coulson, and Bergen}]{michaelov_2023_cloze}
James~A. Michaelov, Seana Coulson, and Benjamin~K. Bergen. 2023.
\newblock \href {https://doi.org/10.1109/TCDS.2022.3176783} {So cloze yet so far: N400 amplitude is better predicted by distributional information than human predictability judgements}.
\newblock \emph{IEEE Transactions on Cognitive and Developmental Systems}, 15(3):1033--1042.

\bibitem[{Nair and Phillips(2026)}]{nair_2026_across}
Sathvik Nair and Colin Phillips. 2026.
\newblock \href {https://arxiv.org/abs/2604.09466} {Across the levels of analysis: Explaining predictive processing in humans requires more than machine-estimated probabilities}.
\newblock \emph{arXiv preprint}, arXiv:2604.09466.

\bibitem[{Nair and Resnik(2023)}]{nair2023words}
Sathvik Nair and Philip Resnik. 2023.
\newblock \href {https://aclanthology.org/2023.findings-emnlp.752/} {Words, subwords, and morphemes: What really matters in the surprisal-reading time relationship?}
\newblock In \emph{Findings of the Association for Computational Linguistics: EMNLP 2023}, pages 11251--11260.

\bibitem[{Nour~Eddine et~al.(2024)Nour~Eddine, Brothers, Wang, Spratling, and Kuperberg}]{nour_eddine_predictive_2024}
Samer Nour~Eddine, Trevor Brothers, Lin Wang, Michael Spratling, and Gina~R. Kuperberg. 2024.
\newblock \href {https://doi.org/10.1016/j.cognition.2024.105755} {A predictive coding model of the {N400}}.
\newblock \emph{Cognition}, 246:105755.

\bibitem[{Oh and Linzen(2025)}]{oh_2025_model}
Byung-Doh Oh and Tal Linzen. 2025.
\newblock \href {https://arxiv.org/abs/2510.05141} {To model human linguistic prediction, make {LLMs} less superhuman}.
\newblock \emph{arXiv preprint}, arXiv:2510.05141v1.

\bibitem[{Oh and Schuler(2023)}]{oh_2023_surprisal}
Byung-Doh Oh and William Schuler. 2023.
\newblock \href {https://doi.org/10.1162/tacl_a_00548} {Why does surprisal from larger {T}ransformer-based language models provide a poorer fit to human reading times?}
\newblock \emph{Transactions of the Association for Computational Linguistics}, 11:336--350.

\bibitem[{Oh and Schuler(2024)}]{oh_2024_leading}
Byung-Doh Oh and William Schuler. 2024.
\newblock \href {https://aclanthology.org/2024.emnlp-main.202} {Leading whitespaces of language models' subword vocabulary pose a confound for calculating word probabilities}.
\newblock In \emph{Proceedings of the 2024 Conference on Empirical Methods in Natural Language Processing}, pages 3464--3472.

\bibitem[{Oh and Schuler(2025)}]{oh_2025_dissociable}
Byung-Doh Oh and William Schuler. 2025.
\newblock \href {https://doi.org/10.1016/j.jml.2025.104645} {Dissociable frequency effects attenuate as large language model surprisal predictors improve}.
\newblock \emph{Journal of Memory and Language}, 143:104645.

\bibitem[{Ohams et~al.(2026)Ohams, Nair, Bhattasali, and Resnik}]{ohams2026predictive}
Chiebuka Ohams, Sathvik Nair, Shohini Bhattasali, and Philip Resnik. 2026.
\newblock \href {https://doi.org/10.1016/j.jml.2025.104705} {A predictive coding model for online sentence processing}.
\newblock \emph{Journal of Memory and Language}, 146:104705.

\bibitem[{Pimentel and Meister(2024)}]{pimentel_2024_compute}
Tiago Pimentel and Clara Meister. 2024.
\newblock \href {https://aclanthology.org/2024.emnlp-main.1020} {How to compute the probability of a word}.
\newblock In \emph{Proceedings of the 2024 Conference on Empirical Methods in Natural Language Processing}, pages 18358--18375.

\bibitem[{Radford et~al.(2019)Radford, Wu, Child, Luan, Amodei, and Sutskever}]{radford_2019_language}
Alec Radford, Jeff Wu, Rewon Child, David Luan, Dario Amodei, and Ilya Sutskever. 2019.
\newblock \href {https://cdn.openai.com/better-language-models/language_models_are_unsupervised_multitask_learners.pdf} {Language models are unsupervised multitask learners}.
\newblock \emph{OpenAI Technical Report}.

\bibitem[{Shain(2024)}]{shain2024word}
Cory Shain. 2024.
\newblock \href {https://doi.org/10.1162/opmi_a_00119} {Word frequency and predictability dissociate in naturalistic reading}.
\newblock \emph{Open Mind}, 8:177--201.

\bibitem[{Shain et~al.(2024)Shain, Meister, Pimentel, Cotterell, and Levy}]{shain_2024_large}
Cory Shain, Clara Meister, Tiago Pimentel, Ryan Cotterell, and Roger Levy. 2024.
\newblock \href {https://doi.org/10.1073/pnas.2307876121} {Large-scale evidence for logarithmic effects of word predictability on reading time}.
\newblock \emph{Proceedings of the National Academy of Sciences}, 121(10):e2307876121.

\bibitem[{Shlegeris et~al.(2024)Shlegeris, Roger, Chan, and McLean}]{shlegeris_2024_language}
Buck Shlegeris, Fabien Roger, Lawrence Chan, and Euan McLean. 2024.
\newblock \href {https://openreview.net/forum?id=RNsnSLdmV7} {Language models are better than humans at next-token prediction}.
\newblock \emph{Transactions on Machine Learning Research}.

\bibitem[{Slaats and Martin(2025)}]{slaats_2025_surprising}
Sophie Slaats and Andrea~E. Martin. 2025.
\newblock \href {https://doi.org/10.1007/s42113-025-00237-9} {What's surprising about surprisal}.
\newblock \emph{Computational Brain \& Behavior}, 8:233--248.

\bibitem[{Smith and Levy(2011)}]{smith_2011_cloze}
Nathaniel~J. Smith and Roger Levy. 2011.
\newblock \href {https://escholarship.org/uc/item/69s3541f} {{Cloze but no cigar: The complex relationship between cloze, corpus, and subjective probabilities in language processing}}.
\newblock In \emph{Proceedings of the Annual Meeting of the Cognitive Science Society}, volume~33, pages 1637--1642.

\bibitem[{Smith and Levy(2013)}]{smith_2013_effect}
Nathaniel~J. Smith and Roger Levy. 2013.
\newblock \href {https://doi.org/10.1016/j.cognition.2013.02.013} {{The effect of word predictability on reading time is logarithmic}}.
\newblock \emph{Cognition}, 128:302--319.

\bibitem[{Speer(2022)}]{robyn_speer_2022_7199437}
Robyn Speer. 2022.
\newblock \href {https://doi.org/10.5281/zenodo.7199437} {rspeer/wordfreq: v3.0}.

\bibitem[{Stanojevi{\'c} et~al.(2023)Stanojevi{\'c}, Brennan, Dunagan, Steedman, and Hale}]{stanojevic2023modeling}
Milo{\v{s}} Stanojevi{\'c}, Jonathan~R. Brennan, Donald Dunagan, Mark Steedman, and John~T. Hale. 2023.
\newblock \href {https://doi.org/10.1111/cogs.13312} {Modeling structure-building in the brain with {CCG} parsing and large language models}.
\newblock \emph{Cognitive Science}, 47(7):e13312.

\bibitem[{Staub(2025)}]{staub2025predictability}
Adrian Staub. 2025.
\newblock \href {https://doi.org/10.1146/annurev-linguistics-011724-121517} {Predictability in language comprehension: Prospects and problems for surprisal}.
\newblock \emph{Annual Review of Linguistics}, 11:17--34.

\bibitem[{Staub et~al.(2015)Staub, Grant, Astheimer, and Cohen}]{staub_2015_influence}
Adrian Staub, Margaret Grant, Lori Astheimer, and Andrew Cohen. 2015.
\newblock \href {https://doi.org/10.1016/j.jml.2015.02.004} {The influence of cloze probability and item constraint on cloze task response time}.
\newblock \emph{Journal of Memory and Language}, 82:1--17.

\bibitem[{Szewczyk and Federmeier(2022)}]{szewczyk_context-based_2022}
Jakub~M. Szewczyk and Kara~D. Federmeier. 2022.
\newblock \href {https://doi.org/10.1016/j.jml.2021.104311} {Context-based facilitation of semantic access follows both logarithmic and linear functions of stimulus probability}.
\newblock \emph{Journal of Memory and Language}, 123:104311.

\bibitem[{Taylor(1953)}]{taylor_1953_cloze}
Wilson~L. Taylor. 1953.
\newblock \href {https://doi.org/10.1177/107769905303000401} {{``Cloze procedure'': A new tool for measuring readability}}.
\newblock \emph{Journalism Quarterly}, 30(4):415--433.

\bibitem[{Timkey et~al.(2025)Timkey, Huang, Oh, Prasad, Arehalli, Linzen, and Dillon}]{timkey_2025_eye}
William Timkey, Kuan-Jung Huang, Byung-Doh Oh, Grusha Prasad, Suhas Arehalli, Tal Linzen, and Brian Dillon. 2025.
\newblock \href {https://osf.io/preprints/psyarxiv/eq2ra_v1} {Eye movements reveal a dissociation between prediction and structural processing in language comprehension}.
\newblock \emph{PsyArXiv}, eq2ra\_v1.

\bibitem[{Vasishth(2006)}]{vasishth_2006_proper}
Shravan Vasishth. 2006.
\newblock \href {https://www.lingexp.uni-tuebingen.de/sfb441/LingEvid2006/abstracts/vasishth.pdf} {{On the proper treatment of spillover in real-time reading studies: Consequences for psycholinguistic theories}}.
\newblock In \emph{Proceedings of the International Conference on Linguistic Evidence}, pages 96--100.

\bibitem[{Wilcox et~al.(2023)Wilcox, Pimentel, Meister, Cotterell, and Levy}]{wilcox_2023_testing}
Ethan~Gotlieb Wilcox, Tiago Pimentel, Clara Meister, Ryan Cotterell, and Roger~P. Levy. 2023.
\newblock \href {https://doi.org/10.1162/tacl_a_00612} {Testing the predictions of surprisal theory in 11 languages}.
\newblock \emph{Transactions of the Association for Computational Linguistics}, 11:1451--1470.

\bibitem[{Xu et~al.(2023)Xu, Chon, Liu, and Futrell}]{xu_2023_linearity}
Weijie Xu, Jason Chon, Tianran Liu, and Richard Futrell. 2023.
\newblock \href {https://aclanthology.org/2023.findings-emnlp.1052} {The linearity of the effect of surprisal on reading times across languages}.
\newblock In \emph{Findings of the Association for Computational Linguistics: EMNLP 2023}, pages 15711--15721.

\end{thebibliography}

\appendix

\section{Number of Observations for Each RT Measure}
\label{app:observations}

Table \ref{tab:observations} outlines the number of observations that were analyzed for each RT measure.

\begin{table*}[ht!]
    \footnotesize
    \centering
    \begin{tabular}{l S[table-format=6.0] S[table-format=5.0] S[table-format=5.0] S[table-format=5.0] S[table-format=5.0] S[table-format=5.0] S[table-format=5.0] S[table-format=5.0] S[table-format=5.0] S[table-format=5.0] S[table-format=5.0]}
    \toprule
    \multicolumn{1}{c}{Measure} & \multicolumn{1}{r}{Total} & \multicolumn{1}{r}{Fold 1} & \multicolumn{1}{r}{Fold 2} & \multicolumn{1}{r}{Fold 3} & \multicolumn{1}{r}{Fold 4} & \multicolumn{1}{r}{Fold 5} & \multicolumn{1}{r}{Fold 6} & \multicolumn{1}{r}{Fold 7} & \multicolumn{1}{r}{Fold 8} & \multicolumn{1}{r}{Fold 9} & \multicolumn{1}{r}{Fold 10} \\ \midrule
    BK21 SPR & 45840 & 4578 & 4601 & 4592 & 4575 & 4573 & 4591 & 4575 & 4574 & 4602 & 4579 \\
    Provo FP & 105958 & 10773 & 10483 & 10505 & 10579 & 10782 & 10512 & 10575 & 10500 & 10575 & 10674 \\
    Provo GP & 105775 & 10749 & 10463 & 10490 & 10564 & 10762 & 10494 & 10565 & 10470 & 10561 & 10657 \\
    UCL SPR & 99865 & 9836 & 10232 & 9923 & 9733 & 10104 & 10015 & 9827 & 10365 & 9858 & 9972 \\
    UCL FP & 41014 & 4130 & 4108 & 4064 & 4147 & 4155 & 4163 & 4061 & 4089 & 4051 & 4046 \\
    UCL GP & 40993 & 4129 & 4101 & 4064 & 4145 & 4152 & 4162 & 4058 & 4087 & 4049 & 4046 \\ \bottomrule
    \end{tabular}
    \caption{The number of observations in each of the 10 folds and their total for each measure.}
    \label{tab:observations}
\end{table*}

\section{Validation of H$_2$ Clustering}
\label{app:cluster_ex}
\begin{figure}
    \centering
    \includegraphics[scale = 0.5]{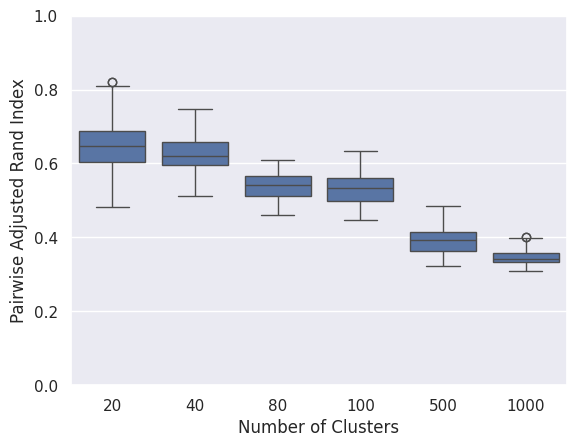}
    \caption{Adjusted Rand Index over all pairs of cluster assignments for different values of $k$ used in H$_2$.}
    \label{fig:rand}
\end{figure}

To determine the stability of the cluster assignments used for H$_2$, we compared the labels assigned to the tokens in GPT2's vocabulary under different clustering settings.
For each value of $k$, we computed a pairwise Adjusted Rand Index \citep{hubert1985comparing} across all five clustering runs (Figure \ref{fig:rand}).
This measures how often different items were assigned to the same cluster, correcting for random chance variation. Although the stability of the clustering decreases as more clusters are used, all results are far above zero, which would be the value of the Adjusted Rand Index under a random baseline.

We also show an illustrative example of the results of clustering GPT2's static token embeddings in Table \ref{tab:cluster_ex}.
This presents select items from five clusters under one of the runs of $k$-means clustering, where $k=80$, reported in Figure \ref{fig:exp2_bymeasure}.
Since this is a qualitative example, we also provide manual characterizations of the clusters.

\begin{table*}[t]
    \centering
    \begin{tabularx}{\textwidth}{>{\hsize=.8\hsize}X>{\hsize=.2\hsize}X} \toprule
    Examples & Characterization \\ \midrule
    \textit{concerned}, \textit{angry}, \textit{worried}, \textit{interested}, \textit{aware}, \textit{proud}, \textit{sick}, \textit{surprised}, \textit{tired}, \textit{convinced} & Emotions \\[3.5ex] 
    \textit{day}, \textit{month}, \textit{weekend}, \textit{annual}, \textit{Nov}, \textit{May}, \textit{afternoon}, \textit{Thursday}, \textit{night}, \textit{February} & Time \\[1ex]
    \textit{leg}, \textit{face}, \textit{foot}, \textit{heart}, \textit{hands}, \textit{wings}, \textit{bones}, \textit{tongue}, \textit{stomach}, \textit{lung} & Body parts \\[1ex] 
    \textit{mother}, \textit{boy}, \textit{father}, \textit{adult}, \textit{twin}, \textit{teenagers}, \textit{grandfathers}, \textit{youth}, \textit{orphan}, \textit{aunt} & Family \\[1ex] 
    \textit{operates}, \textit{extends}, \textit{participates}, \textit{condemns}, \textit{remains}, \textit{uses}, \textit{appears}, \textit{becomes}, \textit{disappears}, \textit{collects} & Agentic verbs \\ \bottomrule
    \end{tabularx}
    \caption{Examples from select $k$-means clusters ($k = 80$) over GPT2's static embeddings and their characterization, used in Run \#5 on the UCL corpus in our evaluation of H$_2$.}
    \label{tab:cluster_ex}
\end{table*}

\section{Complete Results of Experiment 2}
\label{app:exp2_full}
\begin{figure*}
    \centering
    \begin{tabular}{c|c|c}
       $k=20$ & $k=40$ & $k=80$ \\
       \includegraphics[width=0.315\linewidth]{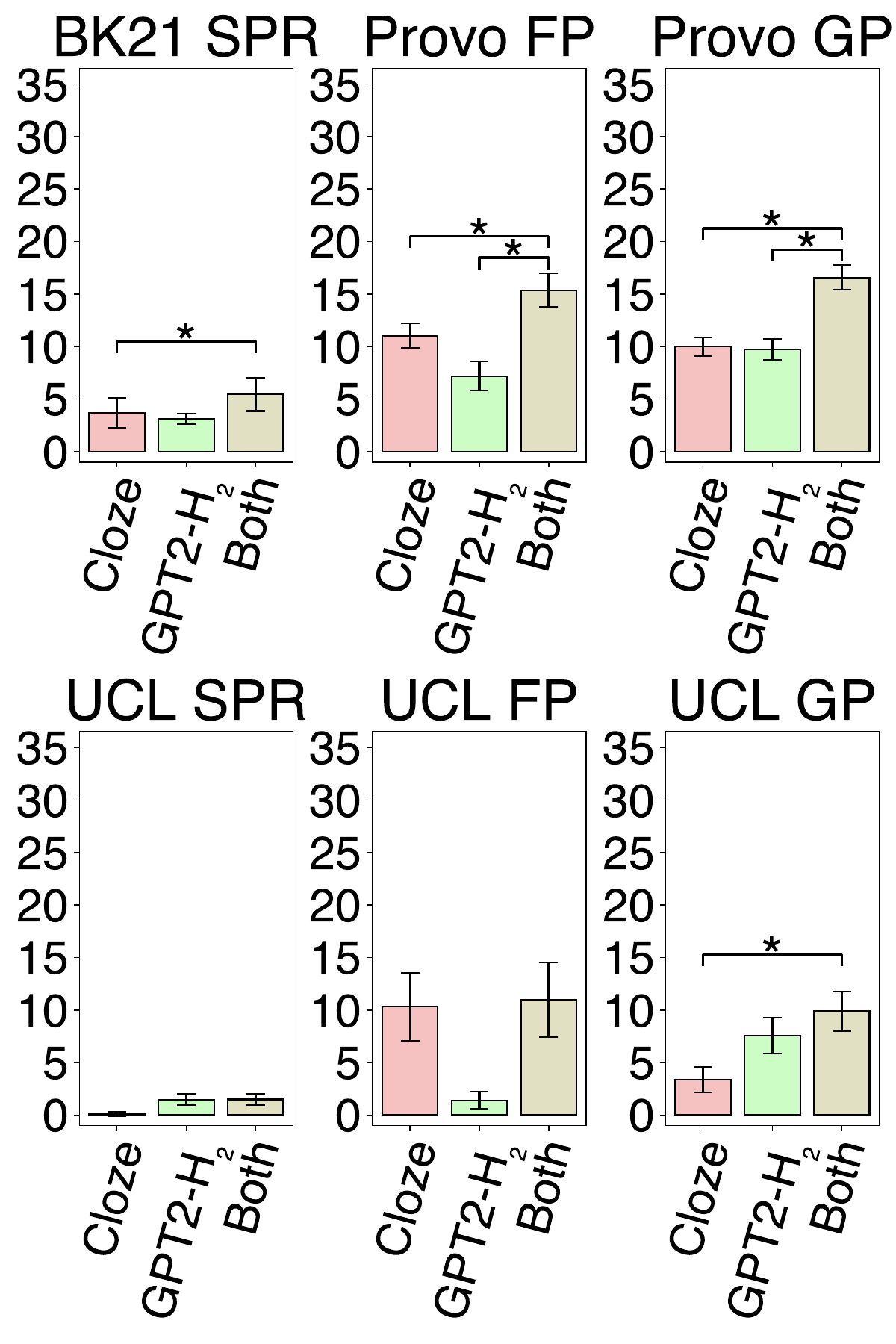} & \includegraphics[width=0.315\linewidth]{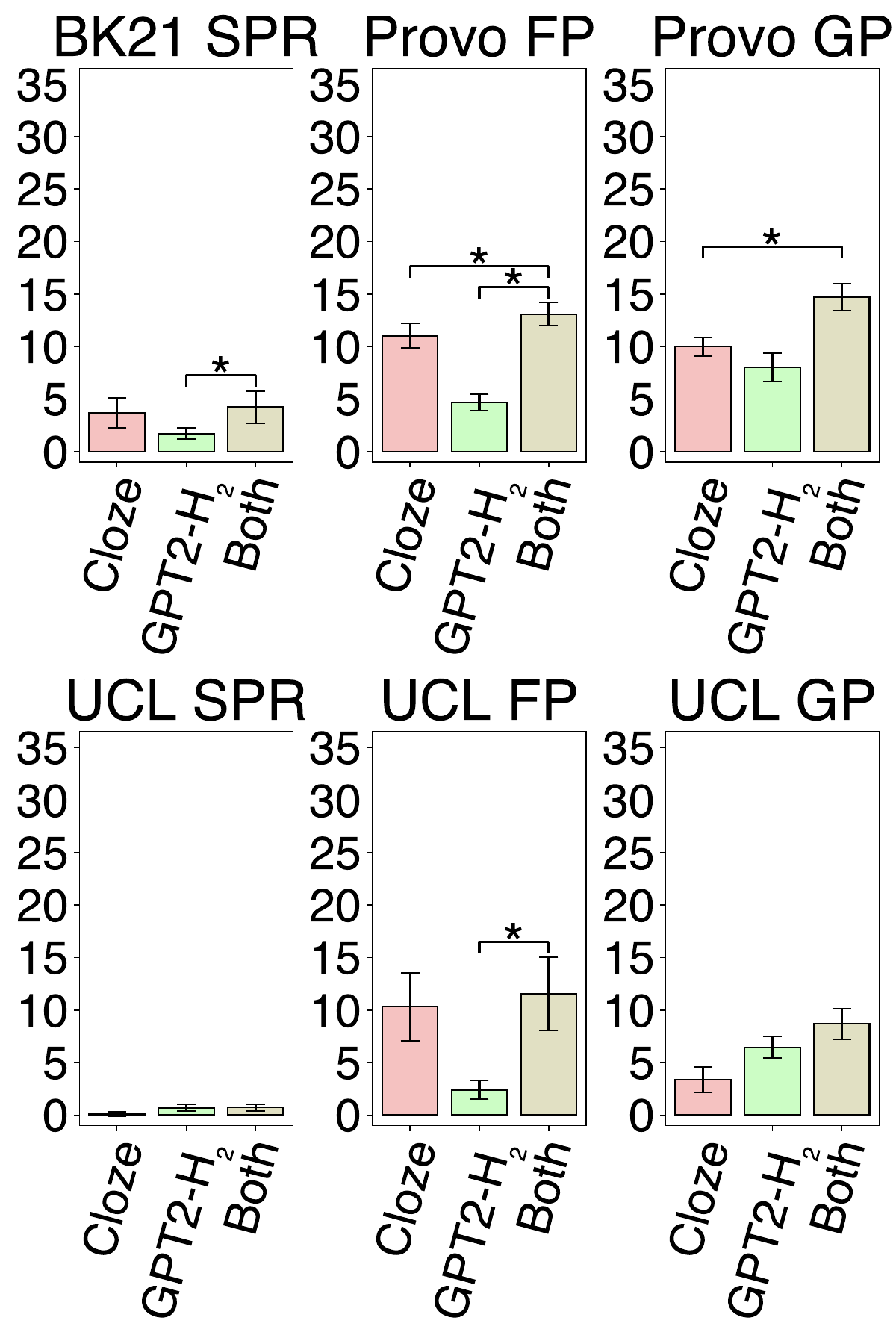} & \includegraphics[width=0.315\linewidth]{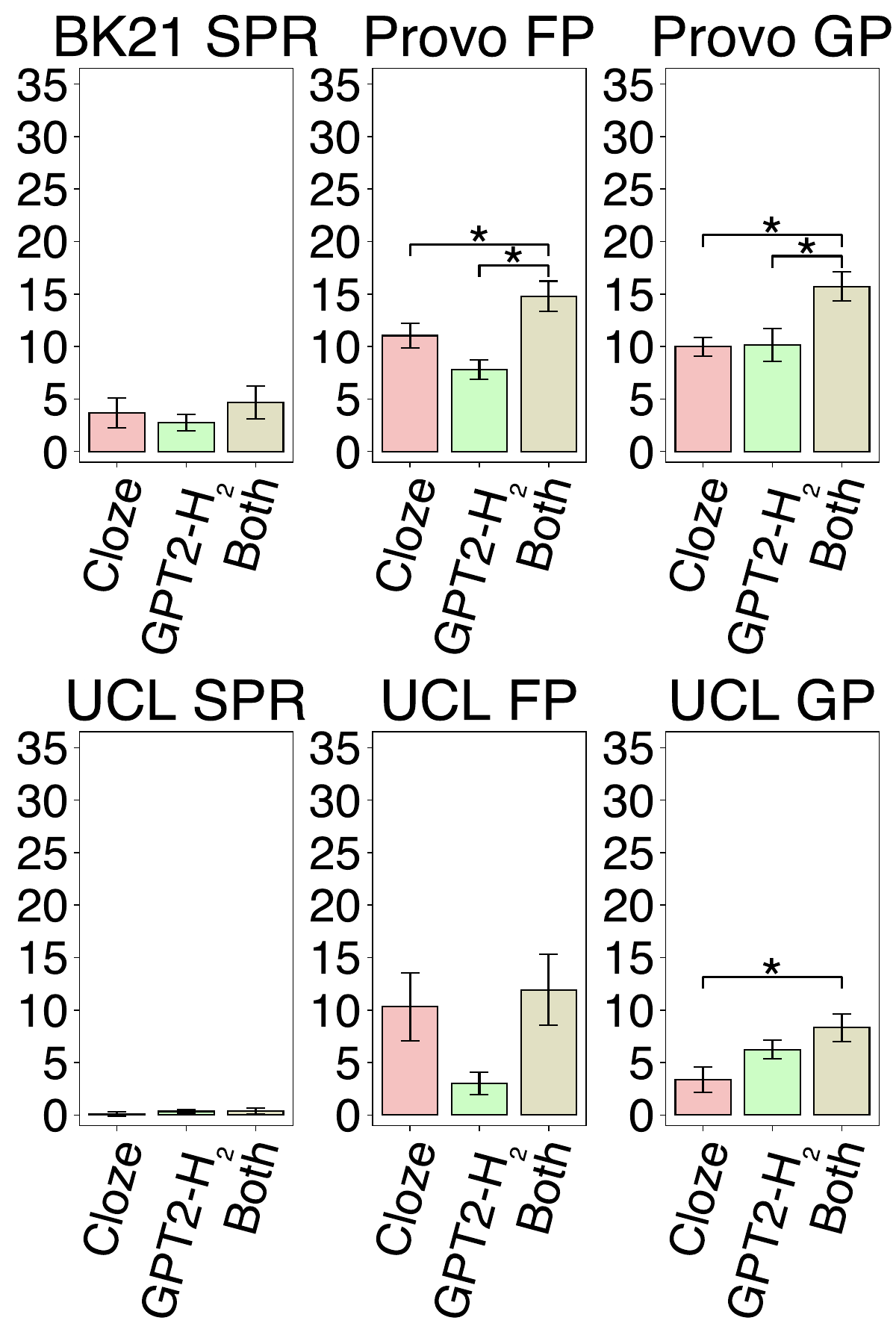} \\
       $k=100$ & $k=500$ & $k=1000$ \\
       \includegraphics[width=0.315\linewidth]{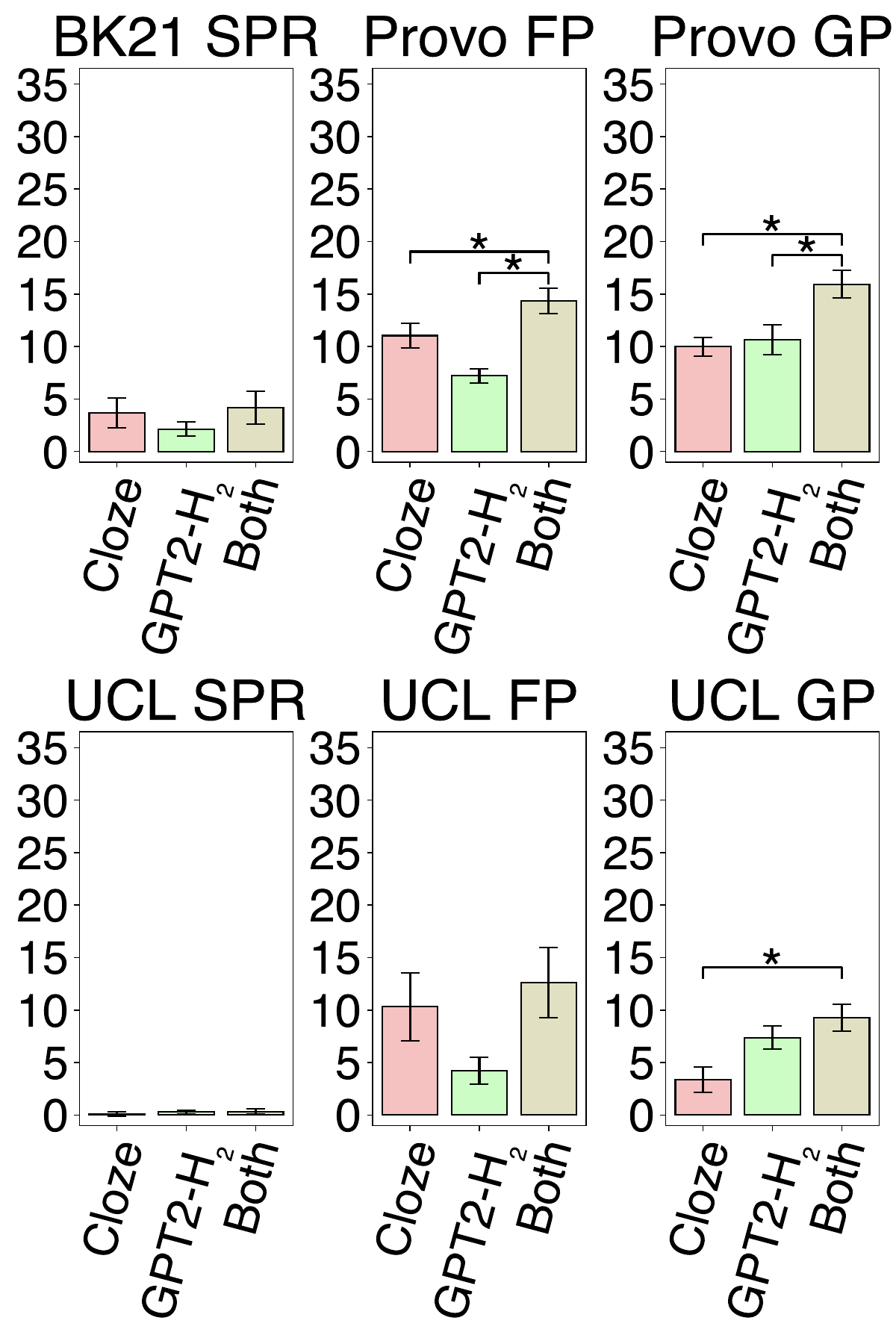} & \includegraphics[width=0.315\linewidth]{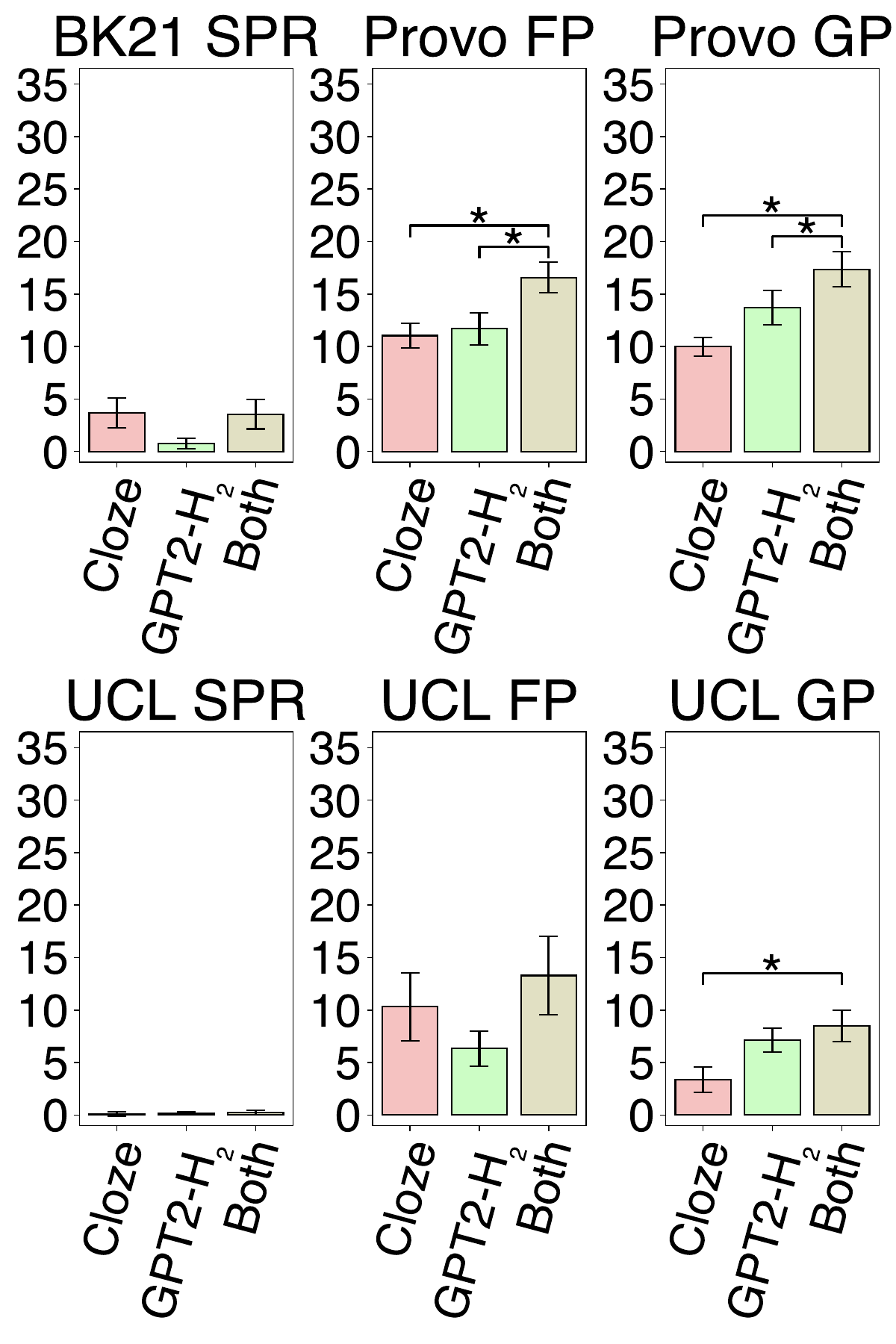} & \includegraphics[width=0.315\linewidth]{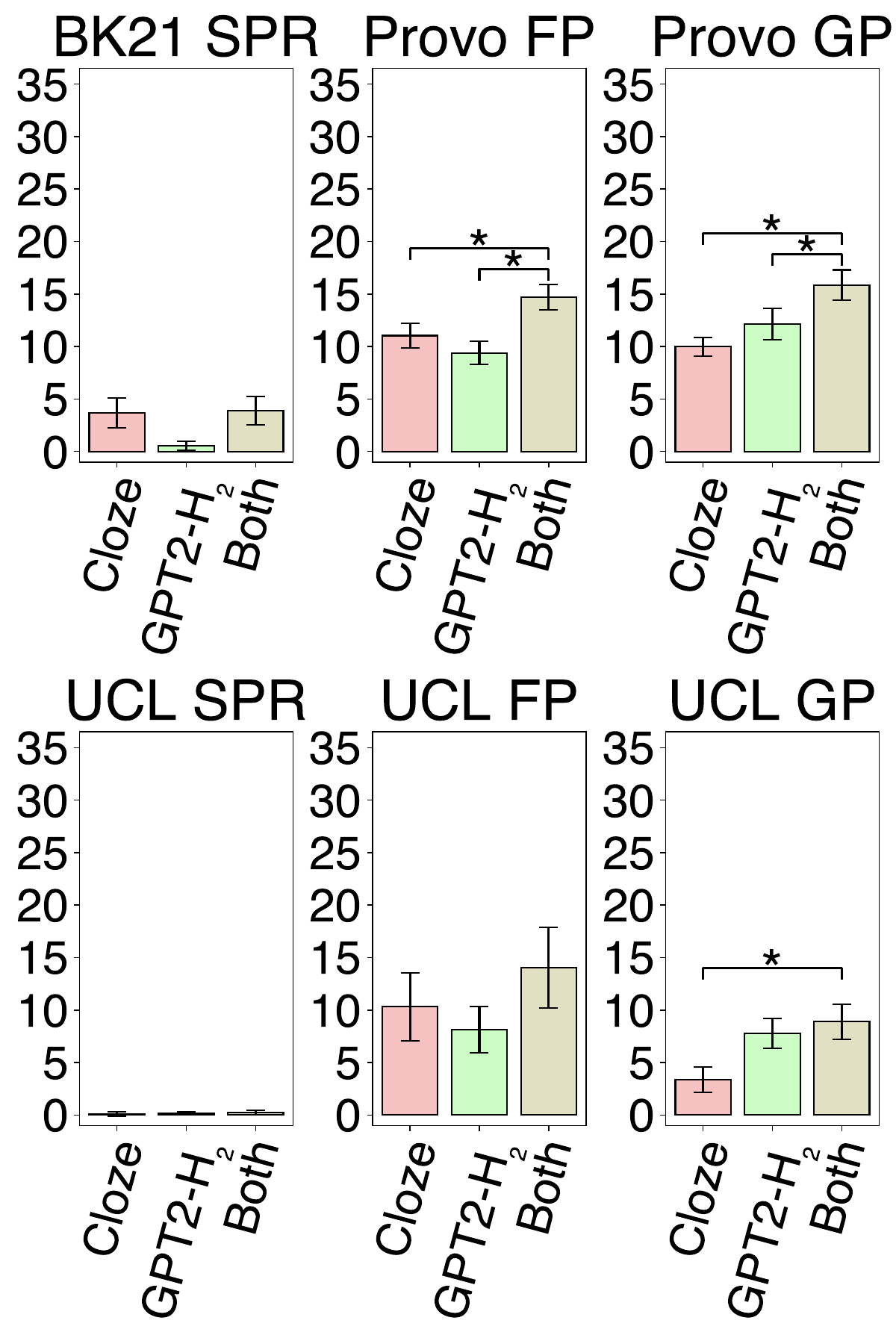}
    \end{tabular}
    \caption{Increase in per-observation log likelihood in $10^{-4}$ nats over the baseline regression models due to including cloze surprisal, GPT-H$_{2}$ surprisal based on different numbers of clusters, and both predictors, averaged over the 10 folds used in cross-validation. Error bars denote one SEM across the 10 folds. Among the two comparisons of interest (Cloze vs.~Both; GPT2-H$_{2}$ vs.~Both), differences that achieve significance at the $0.05$ level by a paired permutation test under a 12-way Bonferroni correction (two comparisons on six measures) are marked with an asterisk.}
    \label{fig:h2_full}
\end{figure*}

Figure \ref{fig:h2_full} shows the performance of GPT2-H$_2$ surprisal for different numbers of clusters used in $k$-means clustering. We report the median results over five runs for each value. The trends in statistical significance remain the same for $k\geq80$, as reported in the main text. 
When $k = 40$, cloze surprisal predicts RTs over and above GPT2-H$_2$ surprisal on BK21 SPR and UCL FP, and there is no significant difference between the two on UCL GP.
When $k = 20$, GPT2-H$_2$ surprisal predicts RTs over and above cloze surprisal on BK21 SPR.

\begin{figure*}
    \centering
    \begin{tabular}{c|c|c}
       Threshold: $10^3$ per billion & Threshold: $10^4$ per billion & Threshold: $10^5$ per billion \\
       \includegraphics[width=0.315\linewidth]{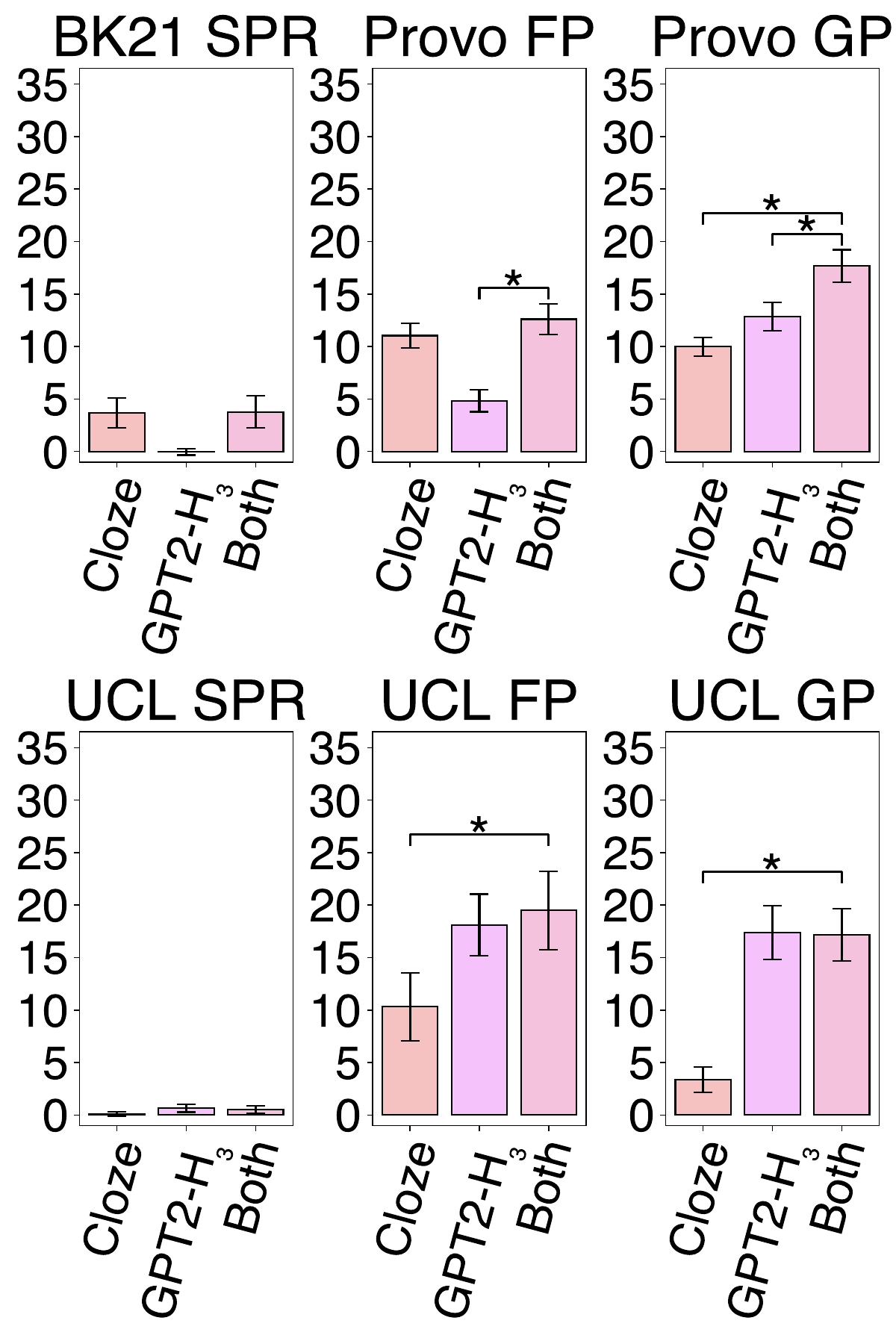} & \includegraphics[width=0.315\linewidth]{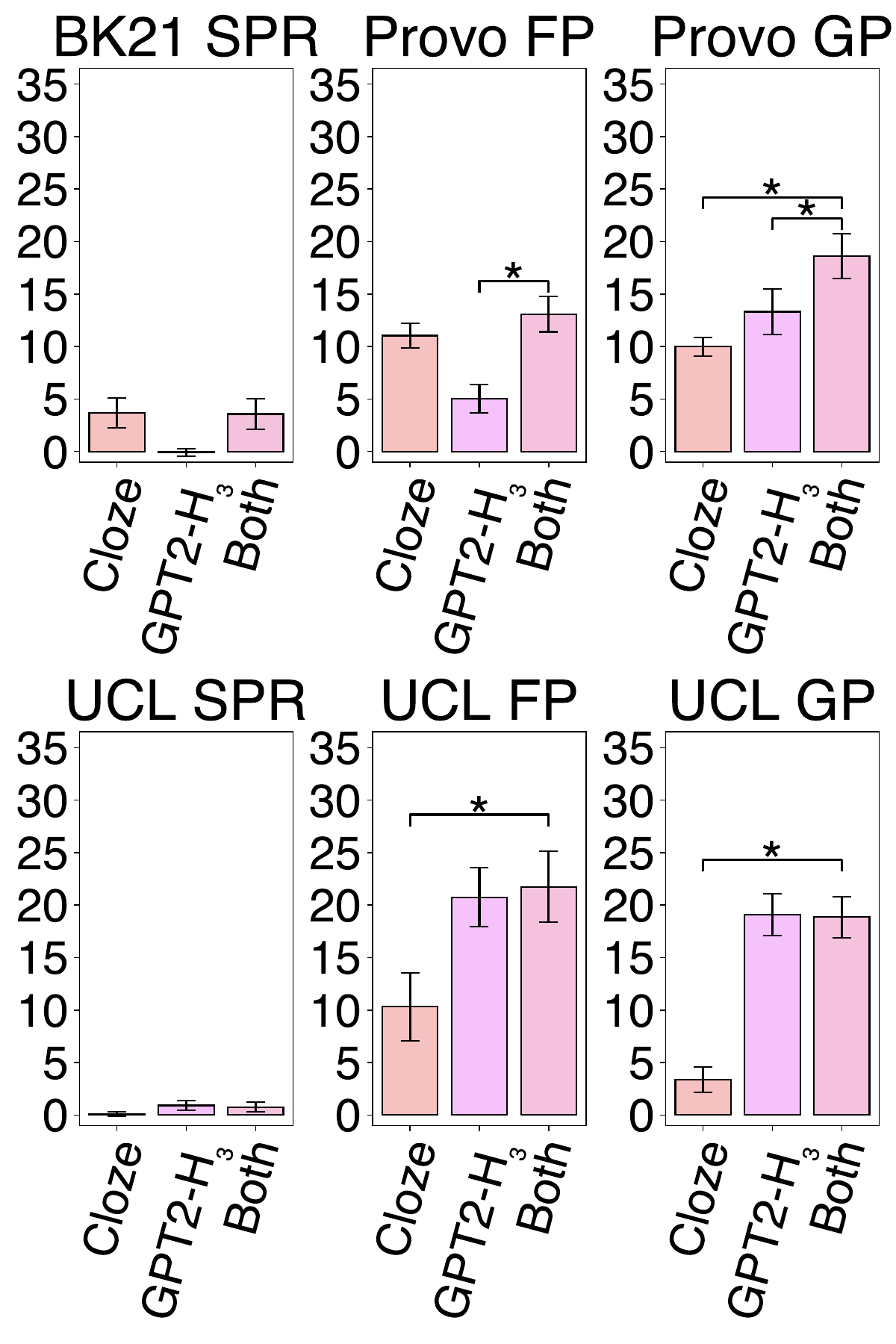} & \includegraphics[width=0.315\linewidth]{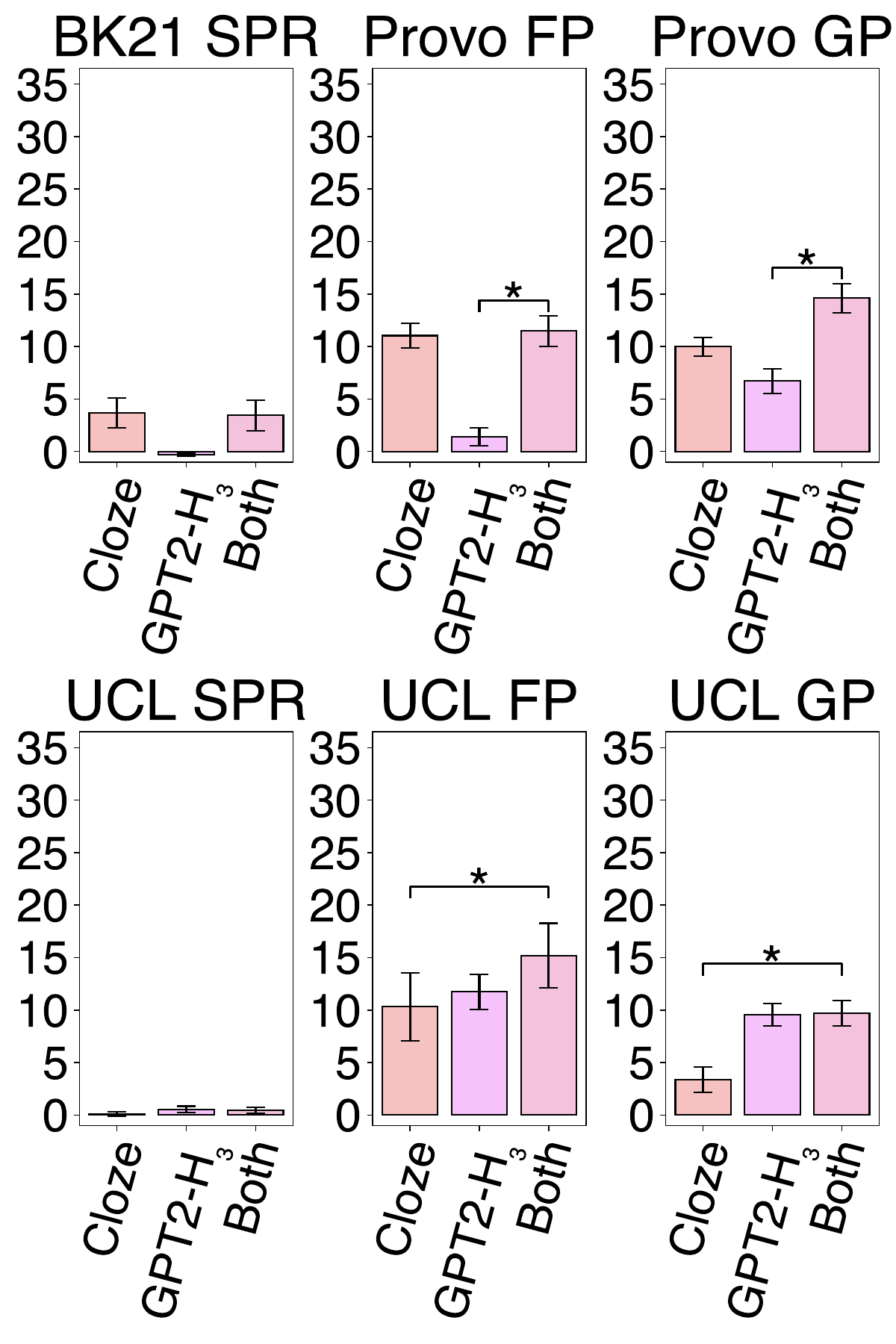}
    \end{tabular}
    \caption{Increase in per-observation log likelihood in $10^{-4}$ nats over the baseline regression models due to including cloze surprisal, GPT-H$_{3}$ surprisal based on different frequency thresholds, and both predictors, averaged over the 10 folds used in cross-validation. Error bars denote one SEM across the 10 folds. Among the two comparisons of interest (Cloze vs.~Both; GPT2-H$_{3}$ vs.~Both), differences that achieve significance at the $0.05$ level by a paired permutation test under a 12-way Bonferroni correction (two comparisons on six measures) are marked with an asterisk.}
    \label{fig:h3_full}
\end{figure*}

Figure \ref{fig:h3_full} shows the performance of GPT2-H$_3$ surprisal for different frequency thresholds. We report results for $10^4$ per billion words (center) in the main text, which shows the same trend as when $10^3$ per billion words is used as the threshold (left). When the threshold is increased to $10^5$ per billion words (right), GPT2-H$_3$ surprisal no longer predicts RTs over and above cloze surprisal on Provo GP.

\begin{figure*}
    \centering
    \begin{tabular}{c|c|c}
       Median & Run \#1 & Run \#2 \\
       \includegraphics[width=0.315\linewidth]{figures/gpt2_h1_slim.pdf} & \includegraphics[width=0.315\linewidth]{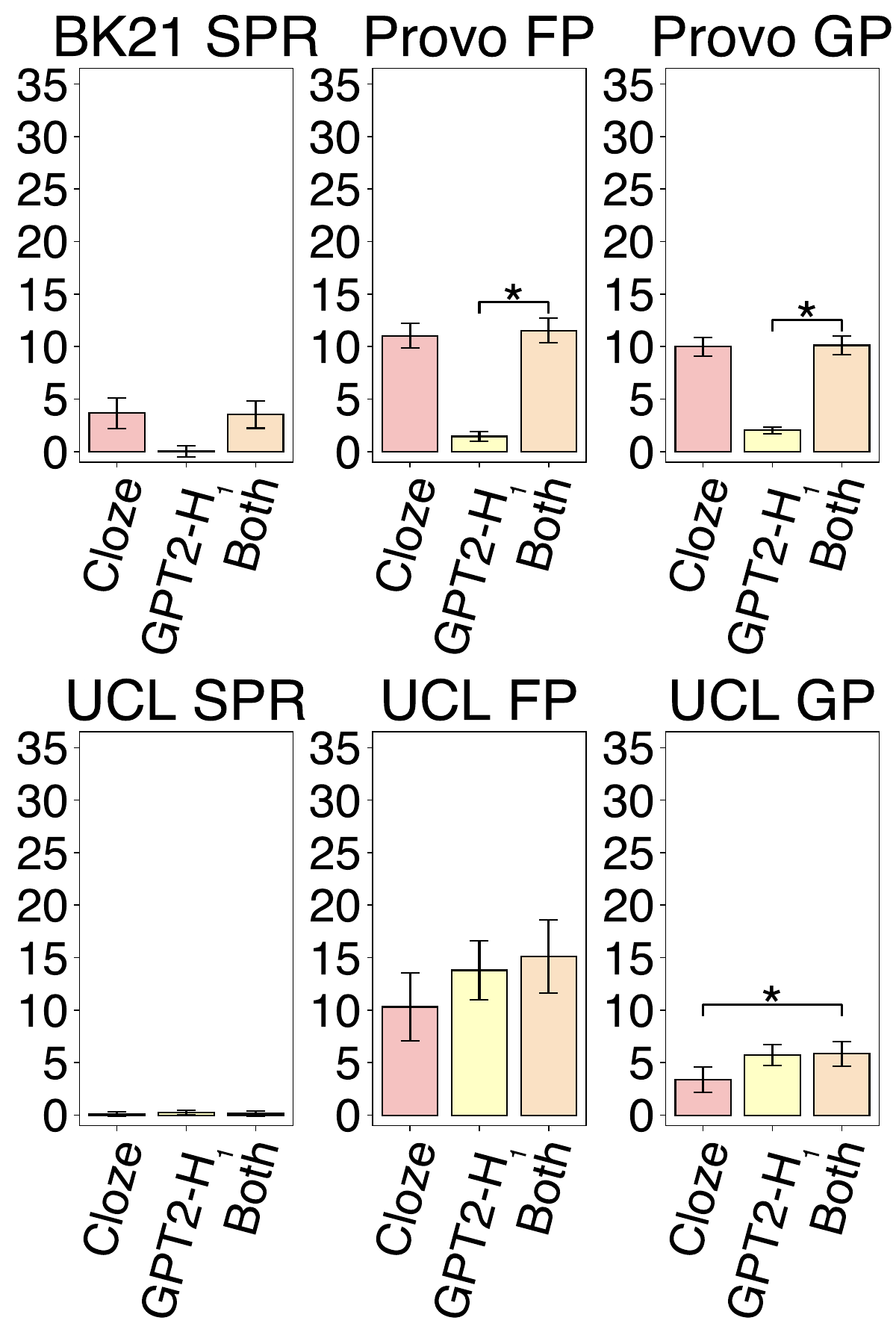} & \includegraphics[width=0.315\linewidth]{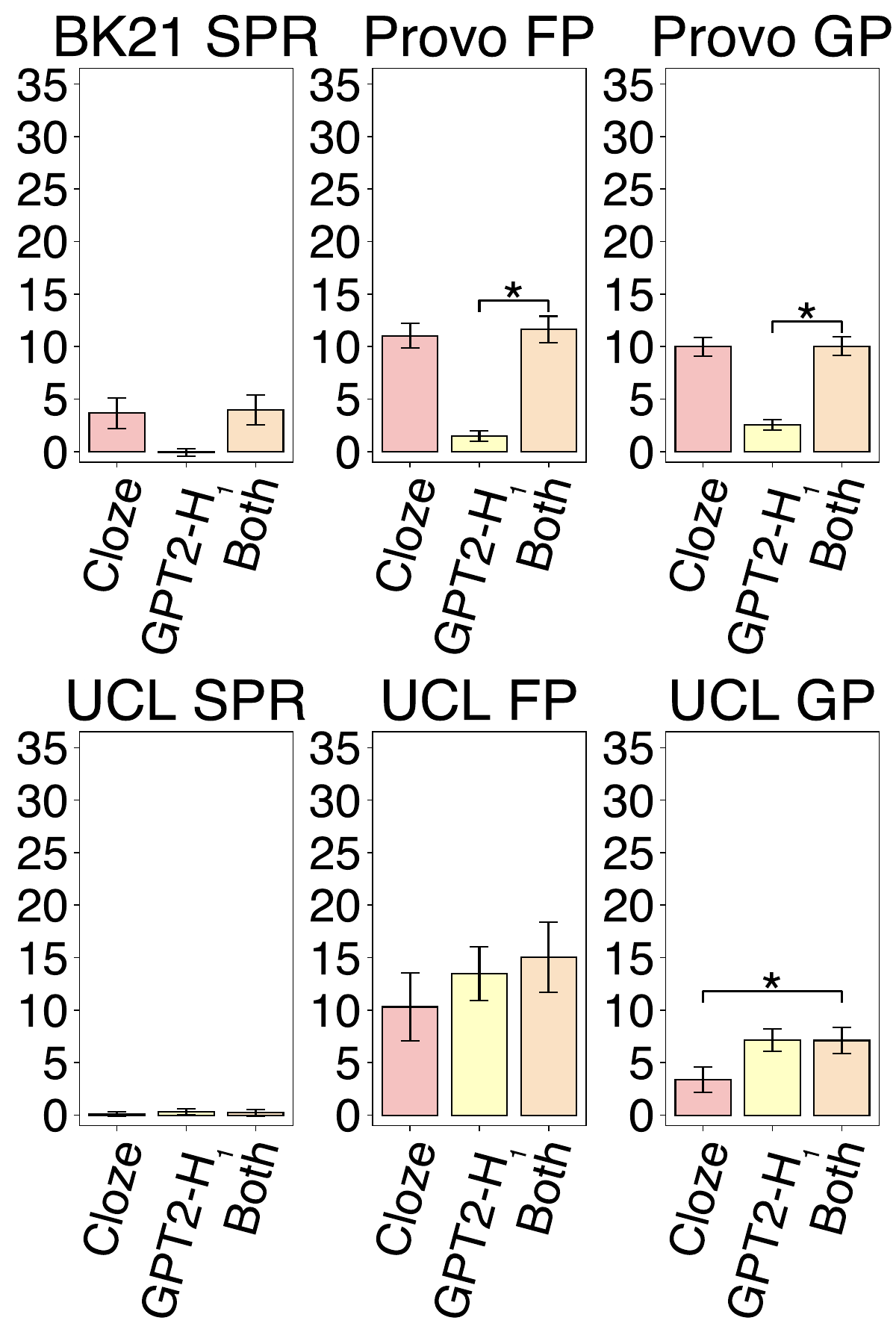} \\
       Run \#3 & Run \#4 & Run \#5 \\
       \includegraphics[width=0.315\linewidth]{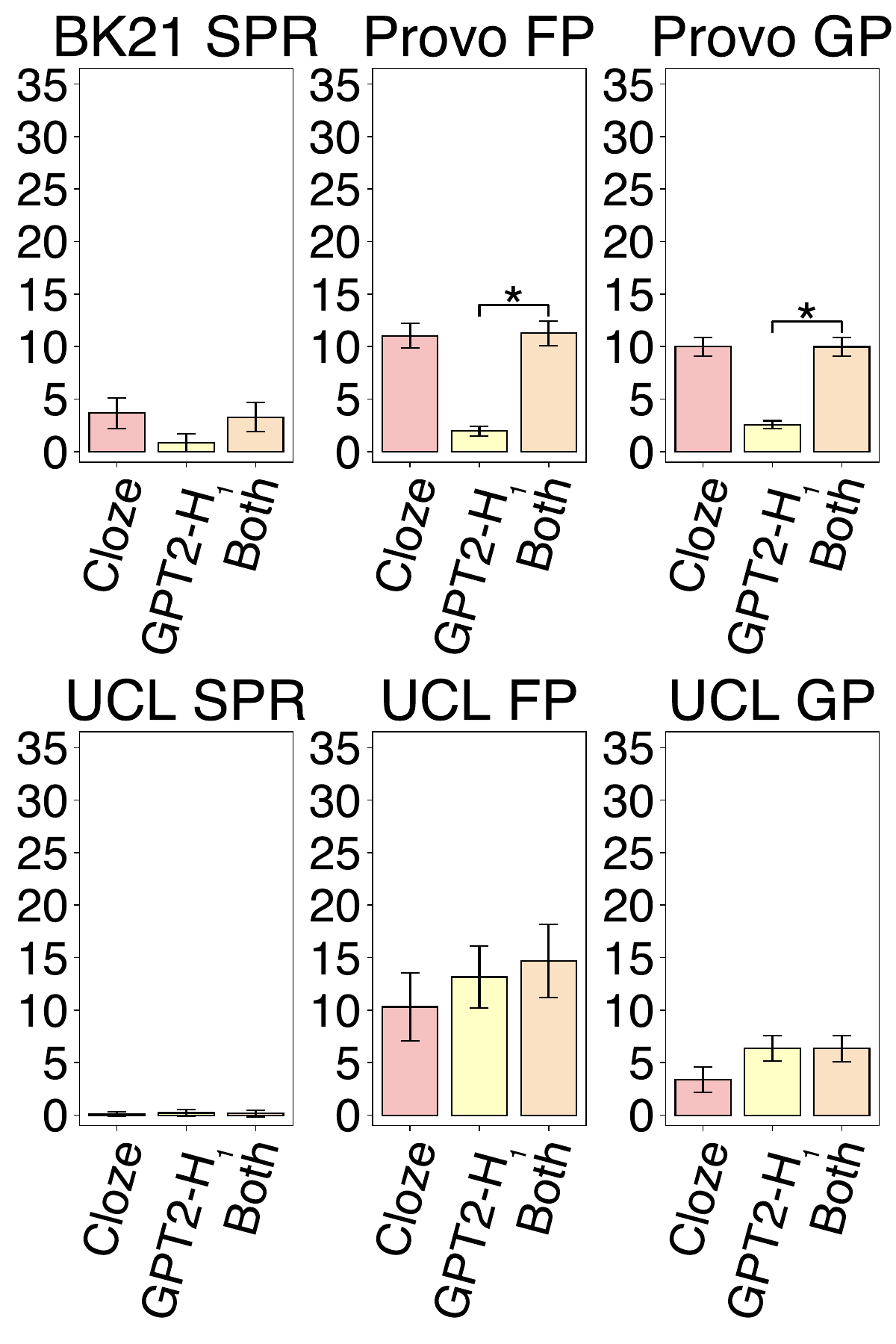} & \includegraphics[width=0.315\linewidth]{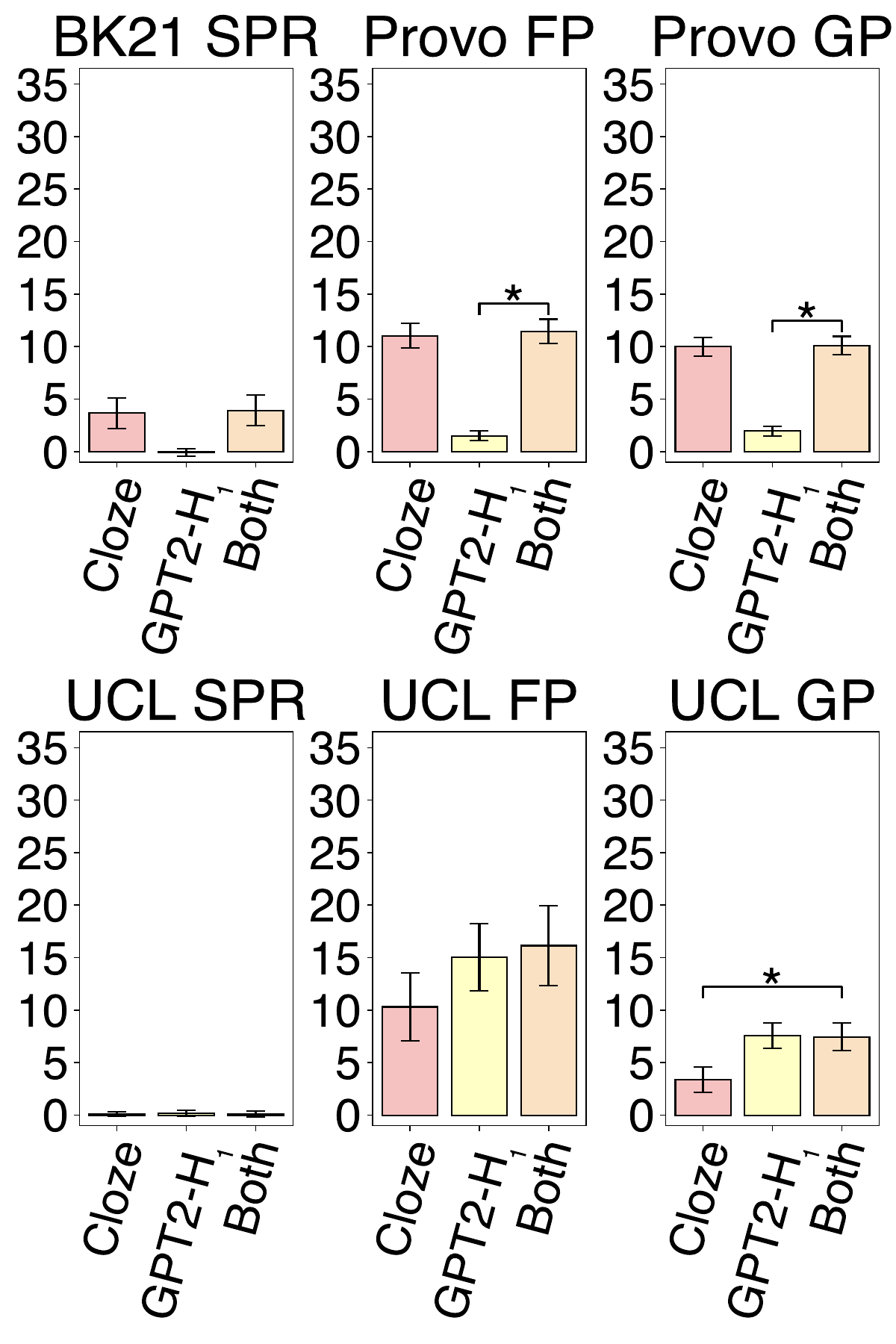} & \includegraphics[width=0.315\linewidth]{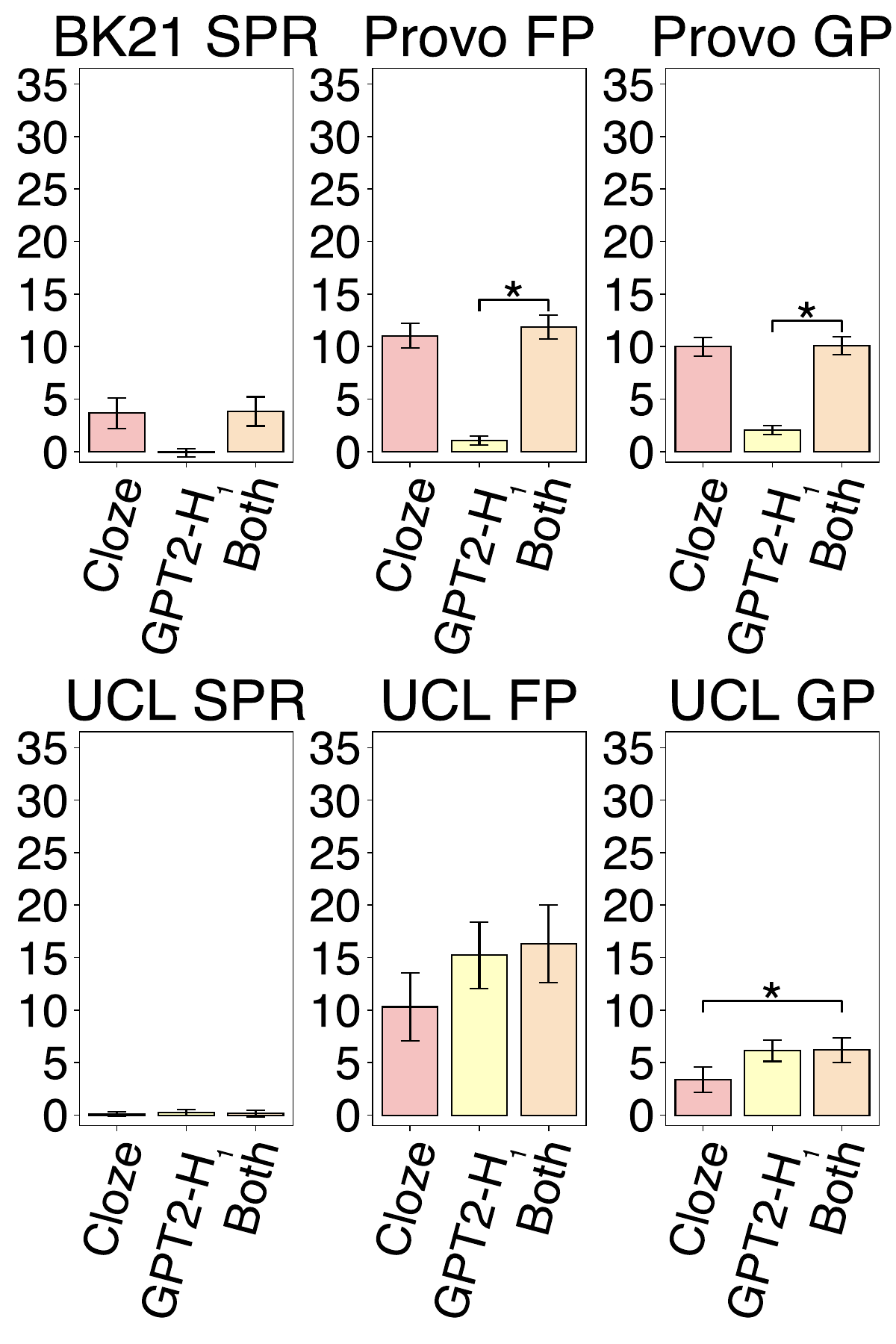}
    \end{tabular}
    \caption{Increase in per-observation log likelihood in $10^{-4}$ nats over the baseline regression models due to including cloze surprisal, GPT-H$_{1}$ surprisal from each sampling run, and both predictors, averaged over the 10 folds used in cross-validation. Results based on the median log likelihood on each fold are repeated for comparison. Error bars denote one SEM across the 10 folds. Among the two comparisons of interest (Cloze vs.~Both; GPT2-H$_{1}$ vs.~Both), differences that achieve significance at the $0.05$ level by a paired permutation test under a 12-way Bonferroni correction (two comparisons on six measures) are marked with an asterisk.}
    \label{fig:h1_individual}
\end{figure*}

\begin{figure*}
    \centering
    \begin{tabular}{c|c|c}
       Median & Run \#1 & Run \#2 \\
       \includegraphics[width=0.315\linewidth]{figures/gpt2_h2_80_clusters.pdf} & \includegraphics[width=0.315\linewidth]{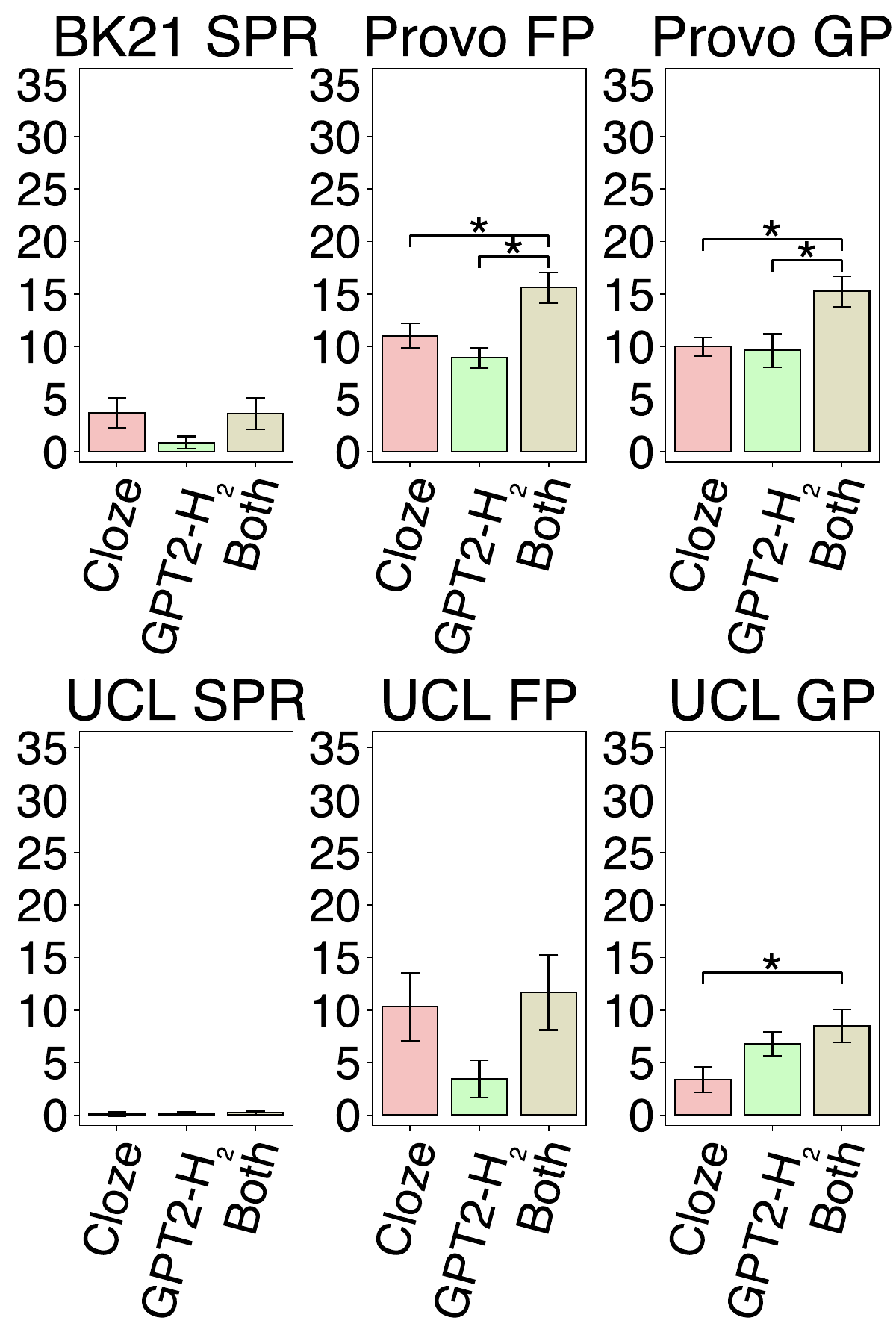} & \includegraphics[width=0.315\linewidth]{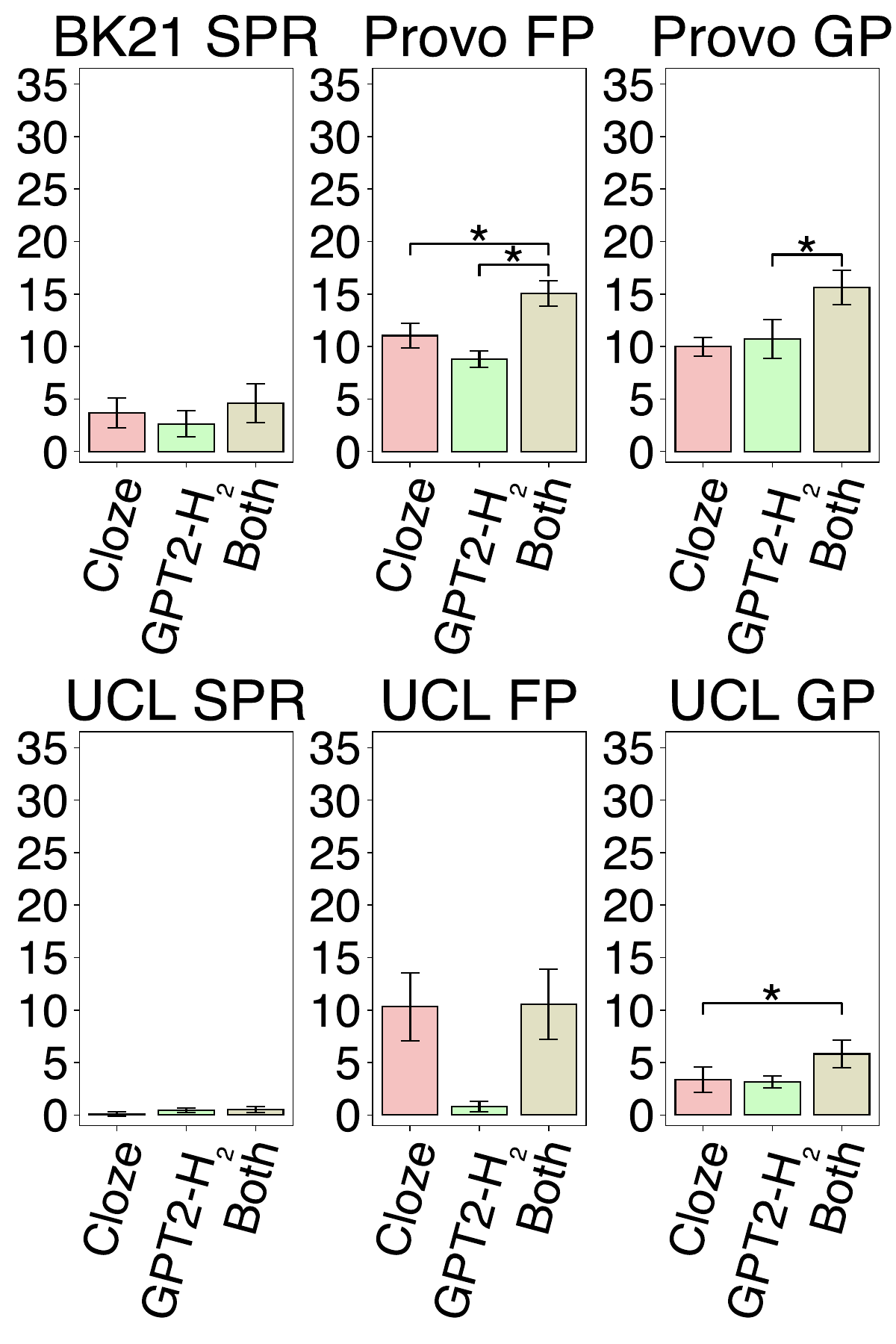} \\
       Run \#3 & Run \#4 & Run \#5 \\
       \includegraphics[width=0.315\linewidth]{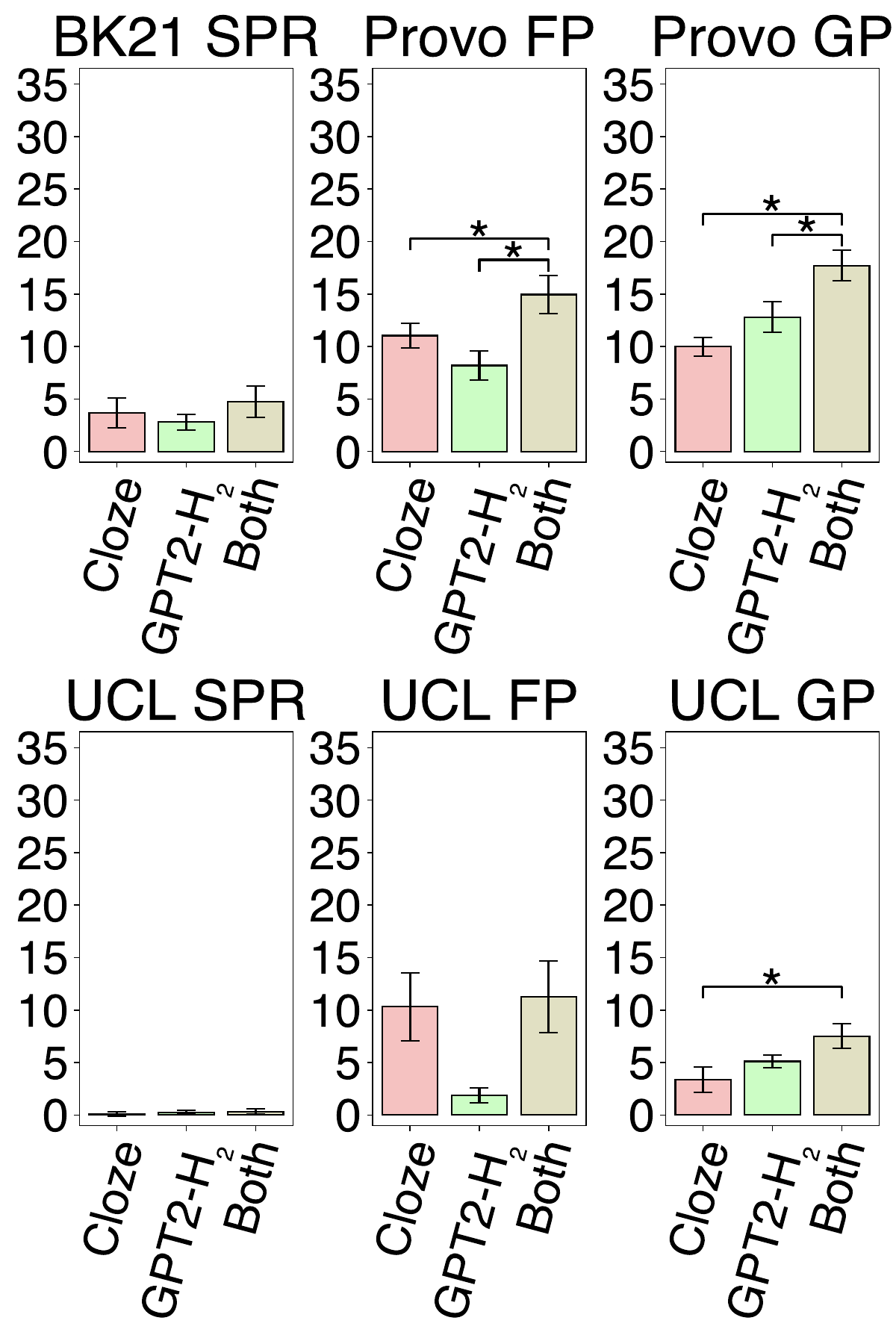} & \includegraphics[width=0.315\linewidth]{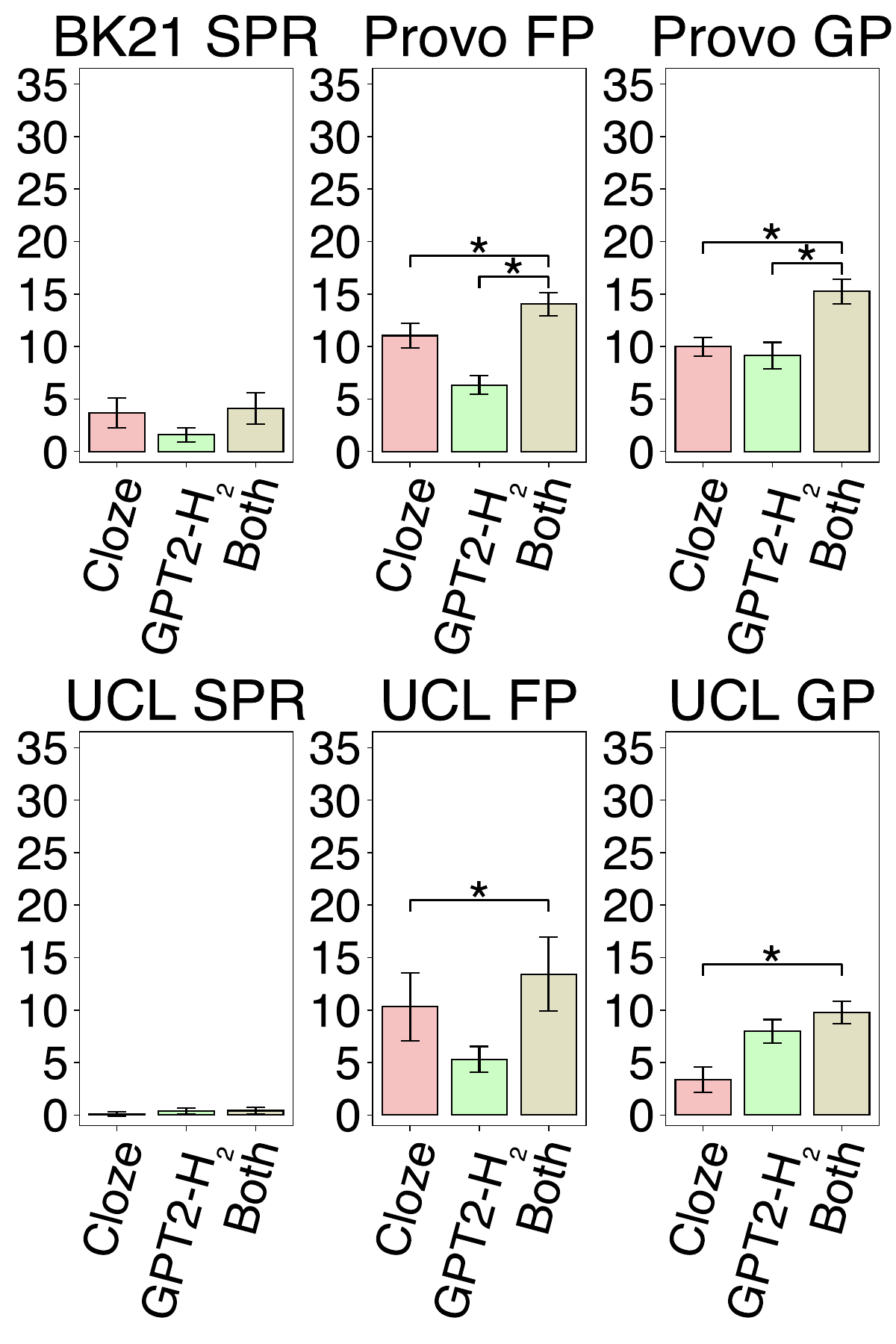} & \includegraphics[width=0.315\linewidth]{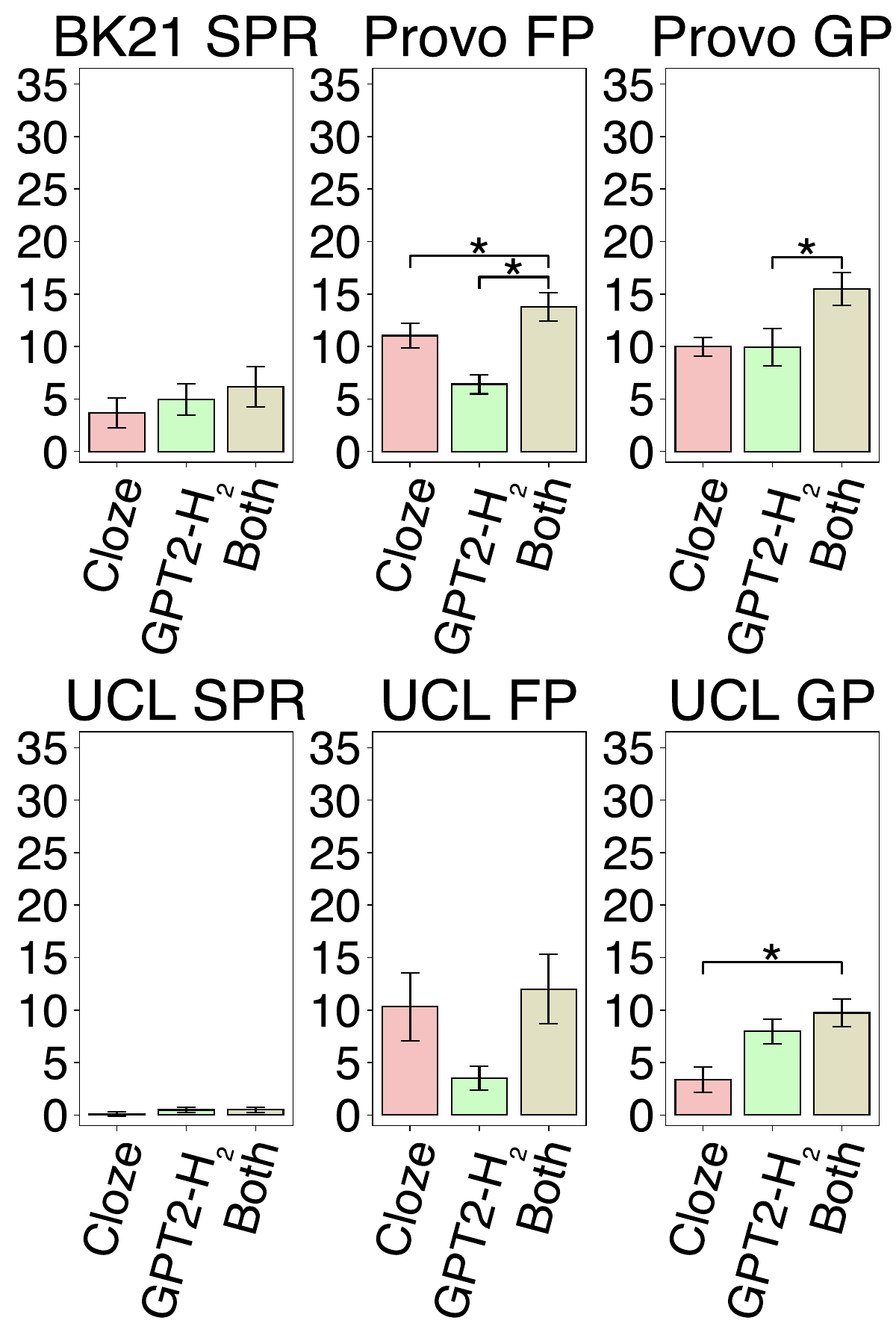}
    \end{tabular}
    \caption{Increase in per-observation log likelihood in $10^{-4}$ nats over the baseline regression models due to including cloze surprisal, GPT-H$_{2}$ surprisal based on 80 clusters from each clustering run, and both predictors, averaged over the 10 folds used in cross-validation. Results based on the median log likelihood on each fold are repeated for comparison. Error bars denote one SEM across the 10 folds. Among the two comparisons of interest (Cloze vs.~Both; GPT2-H$_{2}$ vs.~Both), differences that achieve significance at the $0.05$ level by a paired permutation test under a 12-way Bonferroni correction (two comparisons on six measures) are marked with an asterisk.}
    \label{fig:h2_individual}
\end{figure*}

Figures \ref{fig:h1_individual} and \ref{fig:h2_individual} show results based on individual sampling runs for GPT2-H$_1$ surprisal and clustering runs for GPT2-H$_2$ surprisal respectively.
Across both hypotheses, there is no substantial variability in the magnitude of increase in per-observation log likelihood or the pattern of statistical significance.

\clearpage

\clearpage

\clearpage

\end{document}